\documentclass[11pt]{article}
\usepackage{graphicx, yk} 
\usepackage[margin=2.5cm]{geometry}
\usepackage{float}
\usepackage{amsmath, amsfonts, framed}
\usepackage{amssymb}
\usepackage{tikz}
\usepackage{hyperref}       
\usepackage{textcmds}
\usepackage{algorithm}
\usepackage{algpseudocode}
\let\REQUIRE\Require
\let\ENSURE\Ensure
\let\STATE\State

\allowdisplaybreaks

\newcommand{\IF}[1]{\If{#1}}
\newcommand{\ELSE}{\Else}
\newcommand{\ENDIF}{\EndIf}

\newcommand{\FOR}[1]{\For{#1}}
\newcommand{\ENDFOR}{\EndFor}

\usepackage{adjustbox}
\usepackage{circuitikz}
\usetikzlibrary{arrows.meta, positioning}
\usepackage{booktabs}
\usepackage[normalem]{ulem}
\usepackage{algpseudocode}
\hypersetup{colorlinks,linkcolor={red},citecolor={blue},urlcolor={red}}  
\tikzset{every picture/.style={line width=0.75pt}} 
\usepackage{natbib}
\setcitestyle{authoryear,open={(},close={)}} 

\providecommand{\keywords}[1]
{
  \small	
  \textbf{\textit{Keywords---}} #1
}

\newcommand{\indep}{\perp\!\!\!\!\perp} 

\title{\bf Causal Imitation Learning Under Measurement Error and Distribution Shift}
\author{Shi Bo and AmirEmad Ghassami\\
Department of Mathematics and Statistics, Boston University}
\date{}

\begin{document}

\maketitle

\begin{abstract}
We study offline imitation learning (IL) when part of the decision-relevant state is observed only through noisy measurements and the distribution may change between training and deployment. Such settings induce spurious state--action correlations, so standard behavioral cloning (BC)---whether conditioning on raw measurements or ignoring them---can converge to systematically biased policies under distribution shift. We propose a general framework for IL under measurement error, inspired by explicitly modeling the causal relationships among the variables, yielding a target that retains a causal interpretation and is robust to distribution shift. Building on ideas from proximal causal inference, we introduce \texttt{CausIL}, which treats noisy state observations as proxy variables, and we provide identification conditions under which the target policy is recoverable from demonstrations without rewards or interactive expert queries. We develop estimators for both discrete and continuous state spaces; for continuous settings, we use an adversarial procedure over RKHS function classes to learn the required parameters. We evaluate \texttt{CausIL} on semi-simulated longitudinal data from the PhysioNet/Computing in Cardiology Challenge 2019 cohort and demonstrate improved robustness to distribution shift compared to BC baselines.
\end{abstract}
\keywords{
Imitation Learning; Measurement Error; Distribution Shift; Causal Inference; Spurious Correlation
}

\section{Introduction}

Imitation learning (IL) aims to learn a decision policy from expert demonstrations, without requiring an explicit reward specification. The simplest and most widely used approach is \emph{behavioral cloning} (BC), which reduces IL to supervised learning by fitting a policy that predicts the expert's action from the agent's observation~\citep{pomerleau1989alvinn}. Beyond BC, inverse reinforcement learning \citep{ng2000irl} and generative adversarial imitation learning \citep{ho2016gail} is proposed. 
An important challenge in many real-world settings is that the true state driving the dynamics and expert decisions is \emph{not directly observed}, and the learner instead receives \emph{measurements corrupted by error}. These errors can include noise or systematic sensor bias.
For example, in clinical decision making, clinicians act based on latent aspects of patient condition (e.g., disease severity, overall physiological stability) that may not be directly recorded, while a learning system observes noisy proxies such as vital signs and lab values that can be affected by device calibration, charting practices, or protocol differences across hospitals. 
Analogously, in robotics and autonomous driving, the physical quantities relevant for decision making (e.g., object pose, road friction, occluded agents) are only available through imperfect sensors and perception pipelines whose errors may drift over time.

In this setting, a natural baseline is still to run BC on the available measurements: treat the measurement vector as the policy input and learn $\pi(a\mid o)$ by supervised regression. This is often a sensible choice when there is \emph{no distribution shift}---that is, when the joint distribution of the latent state and the measurement process remains stable between training and deployment. Under such stability, the learned mapping from measurements to expert actions can generalize in the usual supervised-learning sense, and including the measurements (rather than discarding them) is typically beneficial because they carry information about the latent state.
Distribution shift can be due to the environment and/or sensing pipeline changes across domains. This arises, for instance, when a policy trained in one hospital is deployed in another with different device calibration or measurement protocols, or when a driving policy trained with one camera stack is deployed with different calibration or weather-induced perception errors. 
Under shift, purely observational feature--action correlations can become unreliable; in particular, BC can latch onto spurious predictors that are not causally stable across domains, and more information can even hurt due to causal misidentification~\citep{dehaan2019causal}. 
Related causal formulations study imitation when the learner and expert have mismatched sensory inputs and unobserved variables affect decisions~\citep{zhang2020causal,kumor2021sequential}.

\begin{table*}[t]
\centering
\resizebox{\textwidth}{!}{%
\begin{tabular}{lcccc}
\toprule
\textbf{Paper} 
& \textbf{Temporal Confounding} 
& \textbf{Approach} 
& \textbf{Delayed Effect} 
& \textbf{Off-Policy}\\
\midrule

\cite{zhang2020causal} 
& $\times$ 
& Backdoor / Frontdoor Adjustment 
& None 
& Yes \\

\cite{kumor2021sequential} 
& $\times$ 
& Backdoor Adjustment (Sequential) 
& None 
& Yes \\

\cite{swamy2022causal} 
& $\times$ 
& Instrumental Variable 
& $U_{t-1}\!\to\!A_t$ 
& Yes \\

\cite{swamy2022sequence} 
& $\times$ 
& Bayesian + Moment Matching 
& None 
& No \\

\cite{zeng2025confounded} 
& $\checkmark$ 
& Instrumental Variable
& $U_{t-k}\!\to\!A_t$
& Yes \\

\cite{shao2025unifying} 
& $\checkmark$ 
& Instrumental Variable 
& $U_{t-1}\!\to\!S_t$
& Yes \\

\textbf{Ours} 
& $\checkmark$ 
& Proxy-Based Causal Inference 
& $A_{t-1} \rightarrow U_t$, \textbf{$U_{t-1}\!\to\!A_t$} 
& Yes \\

\bottomrule
\end{tabular}%
}
\caption{
Comparison of causal imitation learning frameworks with unobserved confounders. 
\textbf{Temporal Confounding} indicates whether the method explicitly models hidden variables that evolve over time. 
\textbf{Delayed Effect} lists whether past variables influence future actions.}
\label{tab:comparison1}
\end{table*}

This paper focuses on the regime where \emph{measurement error and distribution shift coexist}. 
We demonstrate that in this regime, standard BC can suffer \emph{systematic bias}---bias that does not disappear even as the trajectory length grows and the amount of demonstration data increases. 
Moreover, this pathology is not avoided by simple design choices: both (i) BC that conditions on the raw measurements and (ii) BC variants that exclude measurements or rely on naive denoising can converge to biased targets under shift, because the learned associations depend on non-invariant measurement mechanisms and shifted visitation distributions.
Motivated by this, we propose an alternative target policy that one should prefer over the standard BC target. 
Our proposal is inspired by explicitly modeling the \emph{causal relationships} among latent state, measurements, and actions. 
Rather than imitating the observational conditional $p(a\mid o)$, we define an \emph{optimal causal imitation policy} that is constructed to be robust to changes in (a) the distribution of latent states and (b) the measurement process by reasoning about how actions affect future variables under intervention.

A key difficulty is that this causal-optimal policy is not always identified from observational demonstrations alone. 
We provide identification results that characterize when it \emph{is} identifiable and derive estimands based on measurement variables, drawing inspiration from the proximal causal inference framework for unobserved confounding~\citep{miao2018identifying,tchetgen2020introduction}.
We then develop estimation strategies for the optimal policy in both discrete and continuous settings. 
For discrete state/measurement spaces, we give an estimator based on solving sample analogues of the identification moment equations. 
For continuous settings, we propose an adversarial estimation approach over RKHS function classes.

\subsection{Related Work}

\textbf{Imitation learning.}
Behavioral cloning casts imitation as supervised learning from observations to actions~\citep{pomerleau1989alvinn}. 
Beyond BC, \emph{inverse reinforcement learning} infers a reward/cost function consistent with demonstrations and then plans under that reward~\citep{ng2000irl}, while occupancy-measure matching approaches such as generative adversarial imitation learning align expert and learner behavior by matching visitation distributions via adversarial training~\citep{ho2016gail}. 
As mentioned earlier, a recurring difficulty across IL methods is distribution shift. We note that distribution shift can arise from (i) \emph{endogenous} shift, where the learner's actions change which states are visited relative to the expert, or (ii) \emph{exogenous} shift, where the environment and/or the sensing (measurement) pipeline changes across domains.
The former is the classical IL setting in which compounding errors lead the learner to visit states that are rare or absent in the expert data, motivating interactive data aggregation methods such as DAgger~\citep{ross2011reduction}.
In this work, our primary focus is the latter---shifts in the latent state distribution and/or measurement process across domains---and we show that, when combined with measurement error, standard BC can exhibit systematic bias that does not vanish with increasing trajectory length.

\textbf{Off-policy evaluation under partial observability.}
A related line of work studies off-policy evaluation (and related identification problems) in partially observed sequential decision processes, including POMDPs with latent confounders, where values/policy performance must be estimated from logged trajectories. 
Recent approaches introduce estimators to identify and estimate values under latent confounding and partial observability~\citep{tennenholtz2020off, shi2022minimax, miao2022off, shi2024off, hong2024model}, and proximal reinforcement learning develops efficient off-policy evaluation methods using proxy-variable ideas in partially observed MDPs~\citep{bennett2024proximalrl}. 
These works focus on value estimation with reward feedback, whereas we study imitation without an explicit reward and target a causal-optimal policy under measurement error and distribution shift.

\textbf{Causal Imitation Learning with Latent Confounding.}
Recent work has increasingly framed imitation learning as a causal inference problem.
Several studies highlight \emph{causal delusion} and related biases in sequential models, where past actions spuriously influence posterior beliefs \citep{de2019causal, ortega2021shaking}.
Interactive deconfounded IL methods correct this bias by treating actions as interventions and rely on on-policy expert feedback \citep{vuorio2022deconfounded, swamy2022sequence}.
Orthogonal offline approaches formulate imitation learning as an instrumental-variable problem to address short-range or temporally correlated confounding \citep{swamy2021moments, zhang2020causal, kumor2021sequential, zeng2025confounded}.
As summarized in Table~\ref{tab:comparison1}, existing causal IL methods rely on graphical adjustment, interactive expert correction, or instrumental variable, and do not account for measurement error in state observations.
Our work fills this gap by leveraging the noisy measurements and deriving a estimation strategy that enable fully offline recovery of the expert policy.
A more detailed discussion of related work is provided in Appendix~\ref{Appendix}.

\begin{figure}[t]
    \centering
    \includegraphics[width=0.75\linewidth]{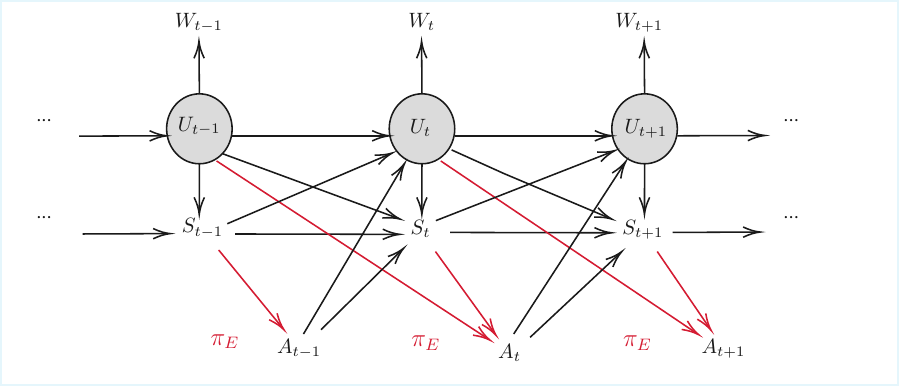}
    \caption{Causal graph for imitation learning under measurement error.}
    \label{fig:proximalME}
\end{figure}

\section{Model Setting and Problem Description}
\label{sec:model}

We study \emph{offline} imitation learning from expert demonstrations when (i) part of the decision-relevant state is latent and affects actions with a one-step delay, and (ii) the learner only observes noisy measurements of this latent state, potentially under distribution shift between training and deployment.

We consider trajectories of length $T$ generated by an expert interacting with an environment. At each time $t=1,\dots,T$, the system contains:
\begin{itemize}
    \item an observed state component $S_t \in \mathcal{S}$ (recorded in the demonstrations);
    \item a latent state component $U_t \in \mathcal{U}$ that is \emph{not} recorded (and may capture unobserved physical/physiological factors, expert beliefs, internal context, etc.);
    \item a measurement (proxy) $W_t \in \mathcal{W}$ that is informative about $U_t$ but may be corrupted by measurement error;
    \item an expert action $A_t \in \mathcal{A}$.
\end{itemize}
The learner observes demonstration data
$\mathcal{D}
=\{(S_{1:T}^{(i)},\,W_{0:T-1}^{(i)},\,A_{1:T}^{(i)})\}_{i=1}^n$,
where $W_{t-1}$ denotes the latest available proxy measurement about the \emph{previous} latent state $U_{t-1}$ when the decision $A_t$ is made.\footnote{This indexing is appropriate in settings where proxy measurements are delayed (e.g., lab tests, batch sensor pipelines), so that $W_{t-1}$ is available at time $t$ while $W_t$ becomes available only after $A_t$ is chosen.}

The expert is assumed to act (possibly approximately) to optimize an unknown objective. Concretely, there exists a reward function $r^\star:\mathcal{S}\times\mathcal{A}\to\mathbb{R}$ such that the expert policy is (approximately) optimal for the induced sequential decision problem, but \emph{neither $r^\star$ nor realized rewards are observed in the demonstrations}. This is the canonical imitation-learning regime: the learner must infer a good policy from state/measurement--action trajectories alone.

A key modeling feature is that the expert's action at time $t$ depends on the \emph{current} observed state $S_t$ and a \emph{delayed} latent state $U_{t-1}$:
\begin{equation}
A_t \sim \pi_E(\cdot \mid S_t, U_{t-1}),
\label{eq:expert_policy}
\end{equation}
where $\pi_E$ is the expert policy and $U_{t-1}$ is unobserved to the learner. This lag captures practical situations where expert decisions depend on internal context that is formed from prior information and persists over time (see Fig.~\ref{fig:proximalME}).
Consider high-speed driving or mobile robotics on varying road surfaces. Let $S_t$ be the current kinematic state (e.g., velocity, yaw rate, distance to obstacles) estimated from onboard sensors. Let $U_t$ be the latent road--tire friction (or wheel--ground contact quality), which is not directly measured and changes slowly. An expert driver (or expert controller) adjusts braking/acceleration at time $t$ using not only the current kinematics $S_t$ but also their \emph{current belief about friction}, which is largely inferred from the \emph{previous} time step via wheel slip and traction signals. This yields a natural delayed dependence $U_{t-1}\rightarrow A_t$. Meanwhile, the learner only observes noisy proxies (e.g., slip ratio, traction-control activation, temperature), summarized as $W_{t-1}$, which can be corrupted and can vary across vehicles or weather conditions.

We model the temporal evolution through a (possibly non-linear) data-generating process in which $U_t$ is persistent and affects future observations, and $W_t$ is a noisy proxy of $U_t$. One convenient abstraction is the causal graph in Fig.~\ref{fig:proximalME}, which implies that $W_{t-1}$ and the recent observed history carry information about $U_{t-1}$, but the learner never observes $U_{t-1}$ directly.
We allow the demonstrations to be collected in one environment (training domain) and the learned policy to be deployed in another (test domain). Distribution shift may occur because the latent dynamics and/or the measurement process changes across domains (e.g., road conditions, sensor calibration, hospital protocols), even when the expert's decision logic in \eqref{eq:expert_policy} is stable.
We consider settings where the variables satisfy the following conditional independences.

\begin{assumption}
\label{assump:proxy_ci}
For each $t\in\{1,\dots,T\}$, let $V_{t-1}:=(S_{t-1},W_{t-1})$. The variables satisfy:
\begin{itemize}
    \item $S_{t-1} \indep A_t \mid (S_t, U_{t-1})$ \quad (past observed state affects $A_t$ only through $(S_t,U_{t-1})$),
    \item $(S_{t-1}, S_t) \indep W_{t-1} \mid U_{t-1}$.
\end{itemize}
\end{assumption}

\subsection{Problem Description}
\label{subsec:problem_description}

We focus on deterministic imitation policies. For a chosen input $X_t$ (e.g., $X_t=S_t$ or $X_t=(S_t,V_{t-1})$), a deterministic policy is a measurable map
\[
\pi:\mathcal{X}\to\mathcal{A},\qquad a_t=\pi(x_t).
\]
In this paper, we treat the expert policy $\pi_E(\cdot\mid S_t,U_{t-1})$ in \eqref{eq:expert_policy} as an \emph{unknown} mechanism and seek to learn a policy from demonstrations only.

A natural learner choice is to fit an observational predictor of $A_t$ from available covariates. We consider two standard behavioral-cloning targets:
{\small
\begin{align*}
\pi_{\mathrm{BC1}}(s_t)
&:= \arg\max_{a\in\mathcal{A}}
\;p(A_t=a \mid S_t=s_t),
\label{eq:bc1_def}
\\
\pi_{\mathrm{BC2}}(s_t,v_{t-1})
&:= \arg\max_{a\in\mathcal{A}}
\;p(A_t=a \mid S_t=s_t, V_{t-1}=v_{t-1}).
\end{align*}
}
Intuitively, $\pi_{\mathrm{BC2}}$ may be more predictive in-sample because $V_{t-1}$ contains information about the delayed latent state $U_{t-1}$.

The following proposition formalizes the connection between deterministic BC and square-loss minimization.

\begin{proposition}[Deterministic BC as a Bayes classifier]\label{prop:bc_l2_det}
Assume $\mathcal{A}=\{1,\dots,K\}$ is finite and let $e(a)\in\mathbb{R}^K$ denote the one-hot encoding of action $a$
(i.e., $e(a)$ has a $1$ in coordinate $a$ and $0$ elsewhere).
Consider the one-hot squared-loss risk over deterministic policies $\pi:\mathcal{X}\to\mathcal{A}$,
\[
\mathcal{R}(\pi)
:=\mathbb{E}_{(X_t,A_t)\sim d_{\pi_E}}\!\left[\left\|e(A_t)-e(\pi(X_t))\right\|_2^2\right].
\]
Then any minimizer $\pi^\star\in\arg\min_{\pi}\mathcal{R}(\pi)$ satisfies, for almost every $x\in\mathcal{X}$,
\[
\pi^\star(x)\in\arg\max_{a\in\mathcal{A}} p(A_t=a\mid X_t=x).
\]
Equivalently, $\mathcal{R}(\pi)=2\,\mathbb{E}\!\left[1-p(A_t=\pi(X_t)\mid X_t)\right]$, so $\pi^\star$ maximizes the conditional success probability pointwise.
\end{proposition}

\begin{remark}
If one instead learns a score function $f:\mathcal{X}\to\mathbb{R}^K$ by minimizing
$\mathbb{E}[\|e(A_t)-f(X_t)\|_2^2]$, the minimizer is $f^\star(x)=\mathbb{E}[e(A_t)\mid X_t=x]$ whose coordinates equal $p(A_t=a\mid X_t=x)$, and the induced policy $\arg\max_a f^\star_a(x)$ coincides with $\pi^\star$ above.
\end{remark}

\paragraph{Concerns regarding using behavioral-cloning.}
Despite being natural observational risk minimizers, $\pi_{\mathrm{BC1}}$ and $\pi_{\mathrm{BC2}}$ fail to directly address two questions that are central in our setting:
\begin{enumerate}
    \item \emph{Causal question:} how would the expert's action change if we were to intervene and set the current state to $S_t=s_t$?
    \item \emph{Deployment question:} how would the learned policy behave when deployed in an environment whose data-generating process differs from that of the training data?
\end{enumerate}
The first question is not answered by observational conditionals such as $p(A_t\mid S_t=s_t,V_{t-1}=v_{t-1})$ because $V_{t-1}$ may encode latent context ($U_{t-1}$) in a way that is correlated with $S_t$, producing non-causal associations.
The second question becomes acute under distribution shift: even when $\pi_{\mathrm{BC2}}$ is highly predictive in the training domain, it can be brittle when the proxy channel or latent dynamics change, because it relies on observational correlations that need not be invariant.

\section{A Causal Imitation Target \label{sec:target}}

The concerns stated in Section \ref{sec:model} motivate an imitation target defined through a \emph{counterfactual intervention}. Let $A_t^{(s)}$ denote the (potential) expert action at time $t$ under the intervention $\mathrm{do}(S_t=s)$, i.e., the action that would be taken if we were to set the current state to $s$ while leaving the rest of the system to evolve naturally.
For discrete actions, we define the causal-optimal imitation policy as
\begin{equation}
\pi_{\mathrm{opt}}(s)
:= \arg\max_{a\in\mathcal{A}} p\!\left(A_t^{(s)}=a\right).
\label{eq:pi_opt_def}
\end{equation}
(For continuous actions, one may analogously define $\pi_{\mathrm{opt}}(s)$ as an interventional conditional mean, but we focus on the discrete form for clarity.)

We next discuss the key properties of $\pi_{\mathrm{opt}}$, demonstrating why it is the right target for our setting.

\textbf{1. Population-aggregated interpretation.} The first key property is regarding the causal interpretation of $\pi_{\mathrm{opt}}$.

\begin{figure*}[t] 
    \centering
    \includegraphics[width=0.3\textwidth]{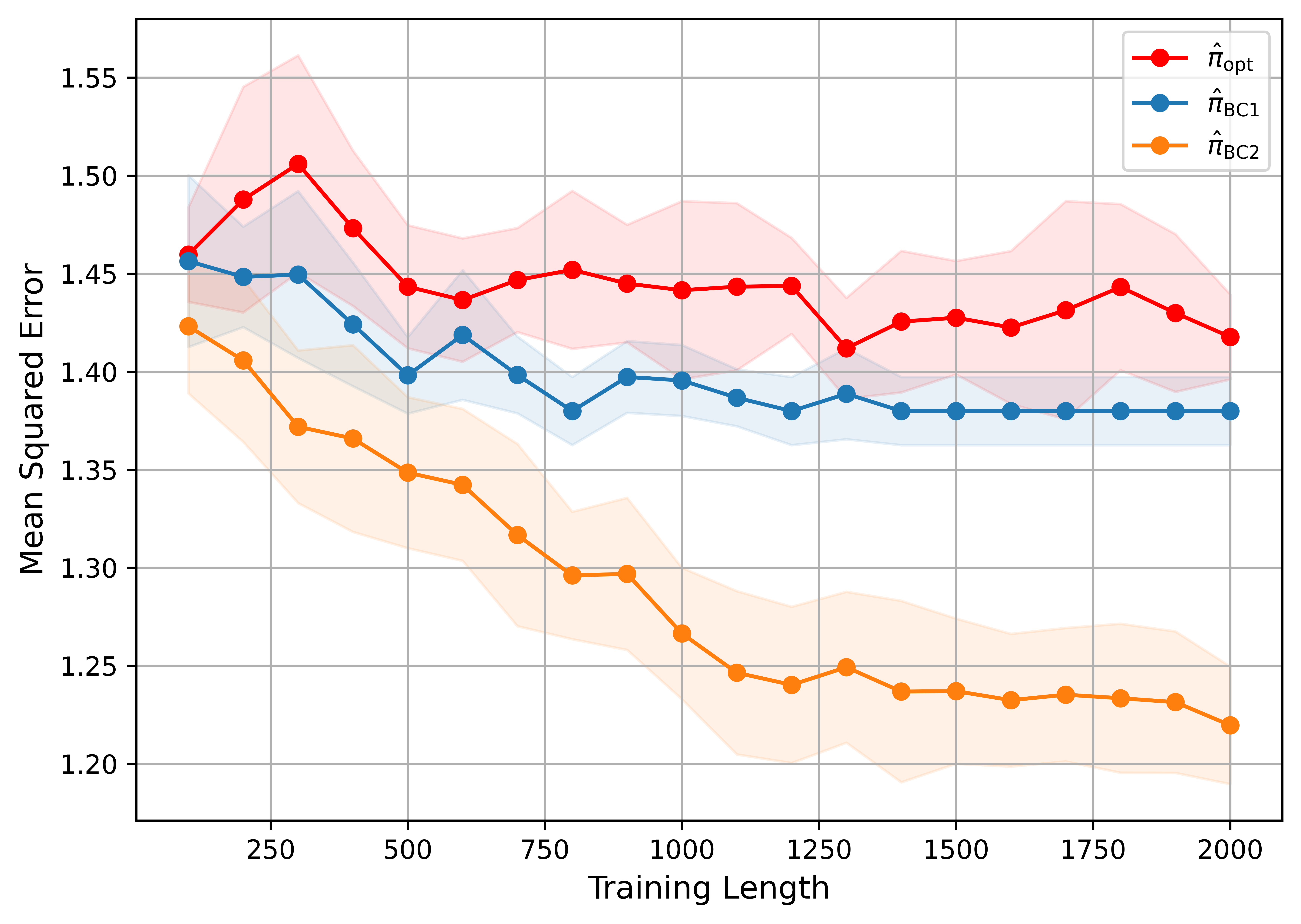}
    \hspace{.2cm}
    \includegraphics[width=0.3\textwidth]{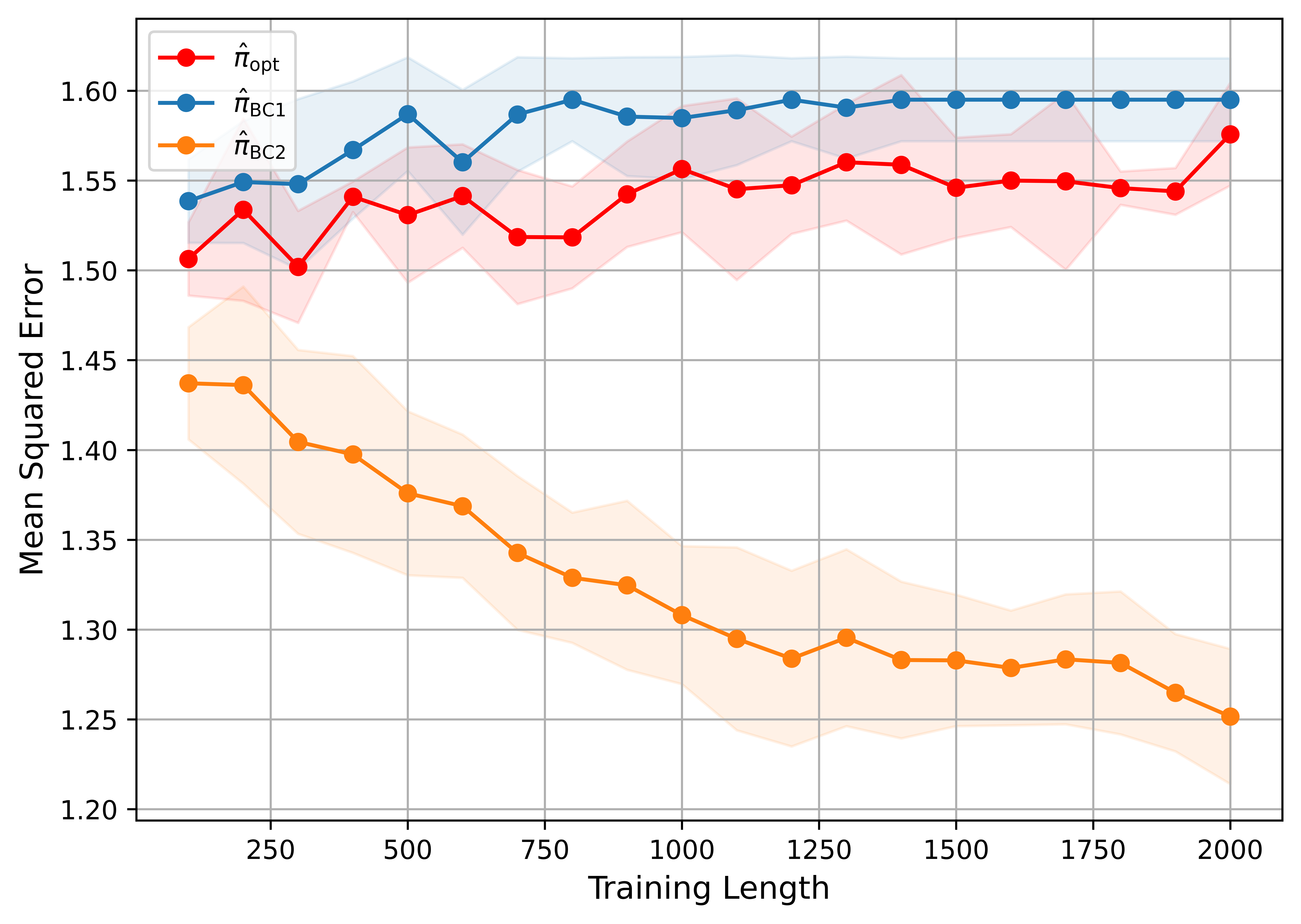}
    \hspace{.2cm}
    \includegraphics[width=0.3\textwidth]{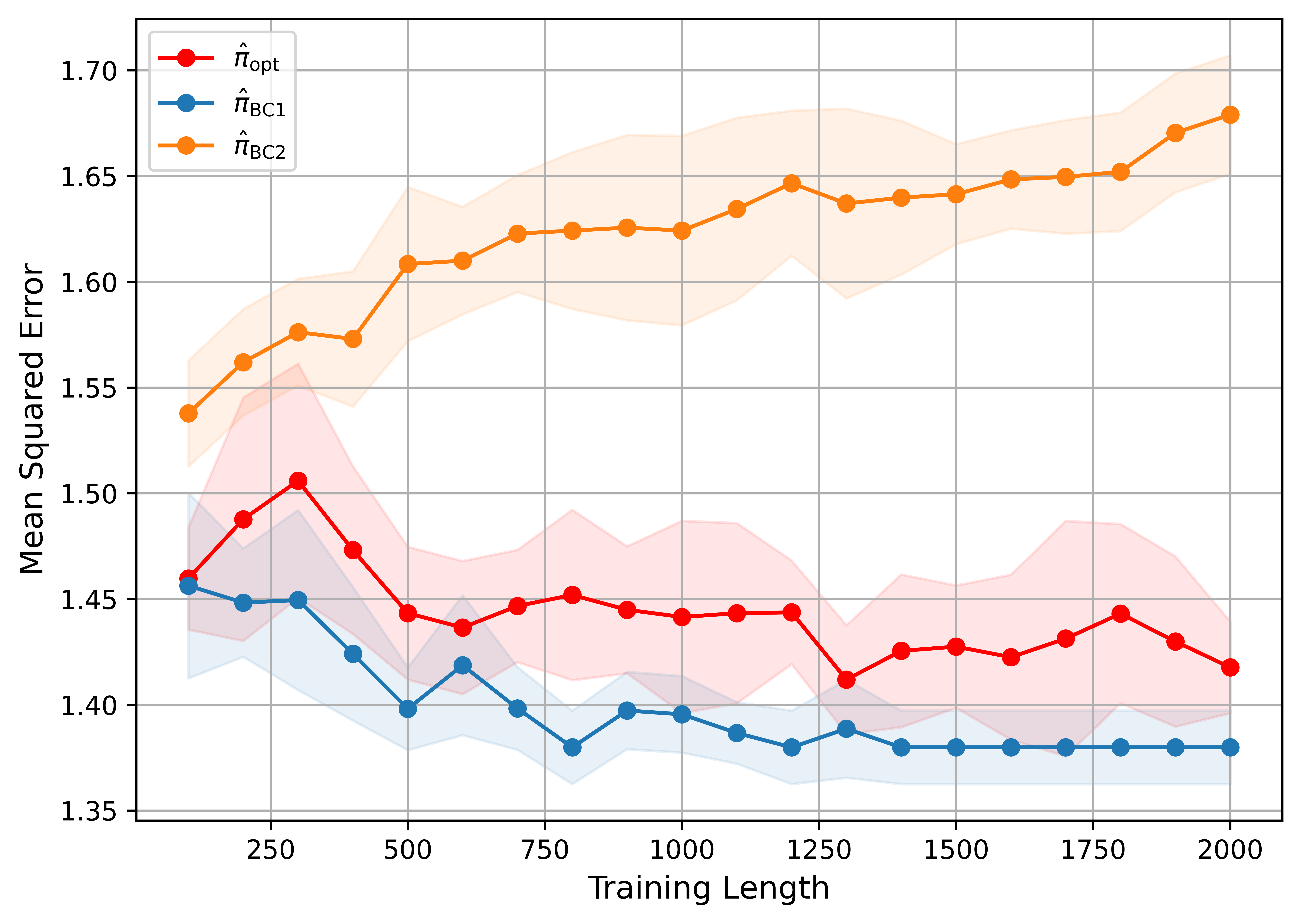}
\caption{
Mean squared error (MSE) of different policies under distributional shift with $|\mathcal A|=4$.
From left to right: (i) no distributional shift; 
(ii) shift in the proxy channel $P(U_{t-1}\mid S_{t})$; 
(iii) simultaneous shifts in $P(W_{t-1}\mid U_{t-1})$.
}
\label{fig:distshift}
\end{figure*}

\begin{proposition}
\label{prop:piopt_core}
Assume $\mathcal{A}$ is finite and the expert follows a deterministic mechanism
\[
A_t=\pi_E(S_t,U_{t-1}).
\]
Assume standard consistency for interventions and temporal ordering, so that under the intervention
$\mathrm{do}(S_t=s)$,
\[
A_t^{(s)}=\pi_E(s,U_{t-1})
\qquad\text{and}\qquad
U_{t-1}^{(s)} = U_{t-1}.
\]
Then for any $s\in\mathcal{S}$ and $a\in\mathcal{A}$,
\begin{align*}
\mathbb{P}\!\left(A_t^{(s)}=a\right)
&= \mathbb{P}\!\left(\pi_E(s,U_{t-1})=a\right) \\
&= \mathbb{E}\!\left[\mathbb{I}\{\pi_E(s,U_{t-1})=a\}\right],
\end{align*}
where the probability/expectation is taken over the (population) distribution of $U_{t-1}$.
Consequently, the causal imitation target
\[
\pi_{\mathrm{opt}}(s)\in\arg\max_{a\in\mathcal{A}} \mathbb{P}\!\left(A_t^{(s)}=a\right)
\]
admits the representation
\[
\pi_{\mathrm{opt}}(s)\in\arg\max_{a\in\mathcal{A}}
\mathbb{E}\!\left[\mathbb{I}\{\pi_E(s,U_{t-1})=a\}\right].
\]
\end{proposition}

\begin{corollary}[Population-aggregated interpretation]
\label{cor:pa}
Define the marginalized (population-aggregated) expert action distribution
\[
\bar{\pi}_E(a\mid s)
:=\mathbb{P}\!\left(\pi_E(s,U_{t-1})=a\right),
\qquad a\in\mathcal{A}.
\]
Then Proposition~\ref{prop:piopt_core} implies
\[
\pi_{\mathrm{opt}}(s)\in\arg\max_{a\in\mathcal{A}} \bar{\pi}_E(a\mid s),
\]
so $\pi_{\mathrm{opt}}(s)$ selects the action that the latent-aware expert mechanism
$\pi_E(s,U_{t-1})$ chooses \emph{most frequently} as $U_{t-1}$ varies in the population.
Equivalently, $\pi_{\mathrm{opt}}(s)$ is Bayes-optimal under $0$--$1$ loss among state-only actions:
\[
\pi_{\mathrm{opt}}(s)\in\arg\min_{a\in\mathcal{A}}
\mathbb{E}\!\left[\mathbb{I}\{\pi_E(s,U_{t-1})\neq a\}\right].
\]
\end{corollary}
\begin{remark}
Because $U_{t-1}$ is unobserved, a policy that conditions only on the current state $s$ cannot, in general, reproduce the latent-aware mapping $\pi_E(s,u)$ uniformly over all latent contexts $u$.
The representation in Proposition~\ref{prop:piopt_core} shows that $\pi_{\mathrm{opt}}(s)$ is defined by a \emph{causal} quantity---the distribution of the expert's action under the intervention $\mathrm{do}(S_t=s)$---rather than an observational conditional.
Corollary \ref{cor:pa} then provides an operational interpretation: $\pi_{\mathrm{opt}}(s)$ is the canonical state-only summary of expert behavior, selecting the action that minimizes disagreement with $\pi_E(s,U_{t-1})$ when the latent context is drawn from its population distribution.
\end{remark}

\textbf{2. Robustness to distribution shift.} The second key property is regarding robustness.
Consider indexing domains by $e\in\{0,1\}$ and write probabilities as $\mathbb{P}_e(\cdot)$.
Assume the expert mechanism $\pi_E(S_t,U_{t-1})$ is invariant across domains, and assume consistency so that under $\mathrm{do}(S_t=s)$, $A_t^{(s)}=\pi_E(s,U_{t-1})$ and $U_{t-1}^{(s)}=U_{t-1}$.

Then for any $s\in\mathcal{S}$ and $a\in\mathcal{A}$,
{\small
\begin{equation}
\mathbb{P}_e(A_t^{(s)}=a)
=\sum_{u\in\mathcal{U}}\mathbb{I}\{\pi_E(s,u)=a\}\,\mathbb{P}_e(U_{t-1}=u),
\label{eq:piopt_domain_dep}
\end{equation}
}
so the target
$\pi_{\mathrm{opt}}^{(e)}(s)\in\arg\max_{a\in\mathcal{A}}\mathbb{P}_e(A_t^{(s)}=a)$
depends on the domain only through the \emph{marginal} distribution $\mathbb{P}_e(U_{t-1})$ (and the invariant mechanism $\pi_E$).

\emph{Robustness relative to $\pi_{\mathrm{BC2}}$.}
The BC2 target is
$\pi_{\mathrm{BC2}}^{(e)}(s,v)\in\arg\max_{a\in\mathcal{A}}\mathbb{P}_e(A_t=a\mid S_t=s,V_{t-1}=v)$,
with $V_{t-1}=(S_{t-1},W_{t-1})$.
Because $W_{t-1}$ is generated through the measurement channel $\mathbb{P}_e(W_{t-1}\mid U_{t-1})$, the posterior
$\mathbb{P}_e(U_{t-1}\mid S_t=s,V_{t-1}=v)$ and hence
$\mathbb{P}_e(A_t\mid S_t,V_{t-1})$ are direct functionals of $\mathbb{P}_e(W_{t-1}\mid U_{t-1})$.
Therefore, a change in the measurement mechanism across domains can change $\pi_{\mathrm{BC2}}^{(e)}$ even when $\pi_E$ is invariant.
In contrast, \eqref{eq:piopt_domain_dep} shows that $\pi_{\mathrm{opt}}^{(e)}$ does \emph{not} involve $\mathbb{P}_e(W_{t-1}\mid U_{t-1})$ at all; in particular, if $\mathbb{P}_0(U_{t-1})=\mathbb{P}_1(U_{t-1})$ and only $\mathbb{P}_e(W_{t-1}\mid U_{t-1})$ changes, then
$\mathbb{P}_0(A_t^{(s)})=\mathbb{P}_1(A_t^{(s)})$ for all $s$ and hence $\pi_{\mathrm{opt}}^{(0)}=\pi_{\mathrm{opt}}^{(1)}$.

\emph{Robustness relative to $\pi_{\mathrm{BC1}}$.}
The BC1 target is
$\pi_{\mathrm{BC1}}^{(e)}(s)\in\arg\max_{a\in\mathcal{A}}\mathbb{P}_e(A_t=a\mid S_t=s)$,
and under $A_t=\pi_E(S_t,U_{t-1})$ we can write
$\mathbb{P}_e(A_t=a\mid S_t=s)
=\sum_{u\in\mathcal{U}}\mathbb{I}\{\pi_E(s,u)=a\}\,\mathbb{P}_e(U_{t-1}=u\mid S_t=s)$.
Thus $\pi_{\mathrm{BC1}}^{(e)}$ depends on $\mathbb{P}_e(U_{t-1}\mid S_t=s)$, which is sensitive to the environment dynamics.
In particular, by Bayes' rule,
$\mathbb{P}_e(U_{t-1}=u\mid S_t=s)\propto \mathbb{P}_e(S_t=s\mid U_{t-1}=u)\,\mathbb{P}_e(U_{t-1}=u)$,
and the conditional likelihood $\mathbb{P}_e(S_t\mid U_{t-1})$ is a functional of the state and latent transition kernels, e.g.
$\mathbb{P}_e(S_t\mid U_t,U_{t-1},S_{t-1},A_{t-1})$ and $\mathbb{P}_e(U_t\mid U_{t-1},S_{t-1},A_{t-1})$,
together with the induced distribution of $(S_{t-1},A_{t-1})$ under the data-generating process.
Therefore, changes in these kernels can alter $\mathbb{P}_e(U_{t-1}\mid S_t=s)$ and hence the BC targets, even if the marginal $\mathbb{P}_e(U_{t-1})$ remains (approximately) the same.

By contrast, \eqref{eq:piopt_domain_dep} shows that $\pi_{\mathrm{opt}}^{(e)}$ depends on the same dynamics only through the marginal $\mathbb{P}_e(U_{t-1})$.
In particular, if a shift changes the kernels
$\mathbb{P}_e(S_t\mid U_t,U_{t-1},S_{t-1},A_{t-1})$ and/or
$\mathbb{P}_e(U_t\mid U_{t-1},S_{t-1},A_{t-1})$ but preserves the marginal of the delayed latent context,
$\mathbb{P}_0(U_{t-1})\approx \mathbb{P}_1(U_{t-1})$, then $\pi_{\mathrm{opt}}$ is approximately invariant.
Formally, for any $s$ and $a$,
\begin{align*}
\left|\mathbb{P}_0(A_t^{(s)}\!=\!a)\!-\!\mathbb{P}_1(A_t^{(s)}\!=\!a)\right|
\!\le\! \|\mathbb{P}_0(U_{t-1})\!-\!\mathbb{P}_1(U_{t-1})\|_1.
\end{align*}
Consequently, whenever the action-probability gap in $\mathbb{P}_e(A_t^{(s)}=\cdot)$ is larger than the induced perturbation above, the $\arg\max$ defining $\pi_{\mathrm{opt}}^{(e)}(s)$ is unchanged across domains.
Finally, it is entirely possible for the transition kernels to change while keeping $\mathbb{P}(U_{t-1})$ approximately fixed; for example, distinct Markov transition kernels can share (exactly or approximately) the same stationary distribution, so changes in dynamics need not imply large changes in the marginal prevalence of $U$.

Figure~\ref{fig:distshift} provides empirical support for the robustness analysis above in terms of mean squared error (MSE) (i.e., compares the actions selected by the learned policy to those of the expert) as defined in Section \ref{sec:experiments}.
When the channel $\mathbb{P}_e(U_{t-1}\mid S_t)$ is perturbed (through a shift changes the kernels
$\mathbb{P}_e(S_t\mid U_t,U_{t-1},S_{t-1},A_{t-1})$), the performance of $\pi_{\mathrm{BC1}}$ degrades noticeably relative to the no-shift case, reflecting its explicit dependence on $\mathbb{P}_e(U_{t-1}\mid S_t=s)$, while $\pi_{\mathrm{opt}}$ remains essentially invariant. 
In contrast, when the measurement mechanism $\mathbb{P}_e(W_{t-1}\mid U_{t-1})$ is altered, $\pi_{\mathrm{BC2}}$ exhibits a clear performance drop, consistent with the fact that its target depends on the posterior $\mathbb{P}_e(U_{t-1}\mid S_t,V_{t-1})$, which is a direct functional of $\mathbb{P}_e(W_{t-1}\mid U_{t-1})$. The details of the data-generating process is provided in Appendix \ref{appendix:simu_setting}.
Overall, the results confirm that $\pi_{\mathrm{BC1}}$ is more sensitive to shifts in $\mathbb{P}_e(U_{t-1}\mid S_t)$, whereas $\pi_{\mathrm{BC2}}$ is vulnerable to changes in the measurement channel, while $\pi_{\mathrm{opt}}$ remains robust across all considered shifts. 

\subsection{Comparison Between $\pi_{\mathrm{opt}}$ and BC}

Under our model, the expert follows a latent-aware mechanism $A_t=\pi_E(S_t,U_{t-1})$, while the learner only observes $(S_t,V_{t-1})$ with $V_{t-1}=(S_{t-1},W_{t-1})$.
Behavioral cloning targets are \emph{observational}: $\pi_{\mathrm{BC1}}(s)$ and $\pi_{\mathrm{BC2}}(s,v)$ maximize conditional action probabilities under the training distribution, which (under determinism) can be written as conditional averages of the latent-aware expert rule with respect to the distribution of the latent context given observed covariates (e.g., $\mathcal{L}(U_{t-1}\mid S_t=s)$ or $\mathcal{L}(U_{t-1}\mid S_t=s,V_{t-1}=v)$).
In contrast, $\pi_{\mathrm{opt}}(s)$ is defined via the interventional response $\mathbb{P}(A_t^{(s)}=\cdot)$ and averages $\pi_E(s,U_{t-1})$ with respect to the \emph{marginal} distribution of $U_{t-1}$.
Since conditioning on $S_t$ or $(S_t,V_{t-1})$ typically changes the latent distribution whenever $U_{t-1}$ is statistically associated with these covariates, the resulting observational and interventional targets generally disagree; moreover, $\pi_{\mathrm{BC2}}(s,v)$ can vary with $v$ and therefore need not coincide with a state-only target.
In the Appendix, we give a simple data-generating process in which $\pi_{\mathrm{BC1}}$, $\pi_{\mathrm{BC2}}$, and $\pi_{\mathrm{opt}}$ yield different decision boundaries even in the population.
They coincide only under strong (and typically unrealistic) conditions, summarized in Proposition~\ref{prop:rare_coincidence}.
\begin{proposition}[Rare coincidence of BC targets and $\pi_{\mathrm{opt}}$]\label{prop:rare_coincidence}
Assume a finite action space $\mathcal{A}$ and a deterministic expert mechanism
$A_t=\pi_E(S_t,U_{t-1})$.
For each $s\in\mathcal{S}$ and $a\in\mathcal{A}$, define the action-induced latent sets
\[
B_a(s):=\{u\in\mathcal{U}:\pi_E(s,u)=a\}.
\]
Then, up to arbitrary tie-breaking,
\begin{align*}
\pi_{\mathrm{opt}}(s)
&\in \arg\max_{a\in\mathcal{A}} \;\mathbb{P}\big(U_{t-1}\in B_a(s)\big),\\
\pi_{\mathrm{BC1}}(s)
&\in \arg\max_{a\in\mathcal{A}} \;\mathbb{P}\big(U_{t-1}\in B_a(s)\mid S_t=s\big),\\
\pi_{\mathrm{BC2}}(s,v)
&\in \arg\max_{a\in\mathcal{A}} \;\mathbb{P}\big(U_{t-1}\in B_a(s)\mid S_t=s,V_{t-1}=v\big).
\end{align*}
Moreover, the following sufficient conditions guarantee coincidence:
\begin{enumerate}
    \item If $U_{t-1}\indep S_t$, then $\pi_{\mathrm{BC1}}(s)=\pi_{\mathrm{opt}}(s)$ for all $s$ (up to ties).
    \item If $U_{t-1}\indep (S_t,V_{t-1})$, then $\pi_{\mathrm{BC2}}(s,v)=\pi_{\mathrm{opt}}(s)$ for all $(s,v)$ (up to ties).
\end{enumerate}
In particular, unless conditioning on $S_t$ (or $(S_t,V_{t-1})$) leaves the induced action probabilities
$\mathbb{P}(U_{t-1}\in B_a(s))$ unchanged---as in the independence cases above, or in degenerate cases where $\pi_E(s,u)$ is (essentially) constant in $u$---the BC targets and $\pi_{\mathrm{opt}}$ are generally different.
\end{proposition}

\section{Identification of the Optimal Policy}
\label{sec:id}

So far, we observed that in IL with measurement error and distribution shift, $\pi_{\mathrm{opt}}$ is a desirable target. We now ask when this policy is identifiable from purely observational expert demonstrations. Our identification strategy is inspired by the proximal causal inference (PCI) framework \citep{miao2018identifying,tchetgen2020introduction} and, in particular, its longitudinal extension for complex longitudinal studies \citep{ying2023proximal}. While PCI is often presented for identifying causal effects on outcome means, our goal is to identify the \emph{full interventional distribution} of the expert's action. The key to proximal identification is access to two proxy variables for $U_{t-1}$. \citep{ying2023proximal} consider a setting with a single ``end of the study'' outcome variable, which is affected by latent confounders at previous time points, and assume each latent confounder has two distinct proxy variables. On the other hand, we have one action variable (which is the variable that we are interested in its potential outcomes) at every time point. We set the lagged state $Z_t:=S_{t-1}$ to play the role of a \emph{treatment-inducing proxy}, and the noisy measurement $W_{t-1}$ to play the role of an \emph{outcome-inducing proxy}. Notably, the \emph{same} state process appears at two time indices: $S_{t-1}$ is used as a proxy for the delayed confounder $U_{t-1}$, while $S_t$ is the variable we conceptually intervene on. This time-indexed ``reuse'' is what makes it possible to identify $p(A_t^{(s)})$ despite unobserved confounding. We require the following core assumptions:

\begin{assumption}[Consistency]
\label{assum:consistency}
For all $t$ and all $s$ in the support of $S_t$, we have 
$S_t=s \implies A_t = A_t^{(s)} \ \text{a.s.}$
\end{assumption}
\begin{assumption}[Positivity]\label{assum:positivity}
For all $t$, all $u$ in the support of $U_{t-1}$, and all $s$ in the support of $S_t$, we have $p(S_t=s\mid U_{t-1}=u) > 0$.
\end{assumption}
\begin{assumption}[Latent exchangeability]\label{assum:latent_exch}
For all $t$ and all $s$, we have
$A_t^{(s)} \;\indep\; S_t \ \big|\ U_{t-1}$.
\end{assumption}
Assumptions~\ref{assum:consistency} and \ref{assum:positivity} are standard.
Assumption~\ref{assum:latent_exch} is a significantly weaker version of the standard conditional exchangeability: it posits that, once one conditions on the \emph{unobserved} confounder $U_{t-1}$, the counterfactual action under $\mathrm{do}(S_t=s)$ is independent of the realized state $S_t$.

\subsection{Discrete Case: Identification via Matrix Inversion}

Assume $Z_t:=S_{t-1}$, $W_{t-1}$, $U_{t-1}$ and $A_t$ are categorical.
For a fixed observed state $s$, define the matrices
$
P_{A\mid Z,s}\in\mathbb{R}^{|\mathcal{A}|\times|\mathcal{Z}|}$,
$P_{W\mid Z,s}\in\mathbb{R}^{|\mathcal{W}|\times|\mathcal{Z}|}$,
and
$P_W\in\mathbb{R}^{|\mathcal{W}|}$,
with entries
\begin{align*}
\left[P_{A\mid Z,s}\right]_{a,z} &:= p(A_t=a\mid S_{t-1}=z, S_t=s),\\
\left[P_{W\mid Z,s}\right]_{w,z} &:= p(W_{t-1}=w\mid S_{t-1}=z, S_t=s),\\
\left[P_W\right]_{w} &:= p(W_{t-1}=w).
\end{align*}

\begin{assumption}
\label{assump:rank_discrete}
Let $k:=|U_{t-1}|$.
Assume $|\mathcal{Z}|\ge k$ and $|\mathcal{W}|\ge k$, and for every $s$ in the support of $S_t$,
the matrix $P_{W\mid Z,s}$ has rank $k$.
Equivalently, for each such $s$ there exist coarsenings
$Z'_t$ of $S_{t-1}$ and $W'_{t-1}$ of $W_{t-1}$ taking values in $\{1,\dots,k\}$ such that
$P_{W'\mid Z',s}$ is square and invertible.
\end{assumption}

Roughly speaking, Assumption~\ref{assump:rank_discrete} requires $S_{t-1}$ and  $W_{t-1}$ to be sufficiently informative about $U_{t-1}$.

\begin{theorem}
\label{thm:identification-discrete}
Under Assumptions~
\ref{assump:proxy_ci},
\ref{assum:consistency} - \ref{assump:rank_discrete}, the interventional action distribution is identified.
In particular, for any $s$ and any coarsenings $(Z'_t,W'_{t-1})$ that make $P_{W'\mid Z',s}$ invertible,
\[
P\!\left(A_t^{(s)}\right)
=
P_{A\mid Z',s}\;
\Big(P_{W'\mid Z',s}\Big)^{-1}\;
P_{W'}\,,
\]
where $P(A_t^{(s)})\in\mathbb{R}^{|\mathcal{A}|}$ denotes the vector with entries
$p(A_t^{(s)}=a)$.
Consequently,
$\pi_{\mathrm{opt}}(s)\;\in\;\arg\max_{a\in\mathcal{A}}\,p\!\left(A_t^{(s)}=a\right)$
is identified for all $s$ (up to ties).
\end{theorem}

\subsection{Continuous Case: Identification via a Confounding Bridge}

We now present the identification result for the continuous case by assuming the following completeness and confounding bridge assumptions which ensures that $S$ and $W$ contain sufficient variation to recover functions of the latent state.

\begin{assumption}[Completeness]
\label{assump:completeness_cont}
For each $s$ in the support of $S_t$, for any square-integrable function $\nu$, if
    $\mathbb{E}\!\left[\nu(U_{t-1})\mid S_{t-1},S_t=s\right]=0$ a.s., then $\nu(U_{t-1})=0$ a.s.
\end{assumption}

\begin{assumption}[Confounding bridge]
\label{assump:bridge}
For each action $a\in\mathcal{A}$, there exists a function $h_a:\mathcal{W}\times\mathcal{S}\to\mathbb{R}$
such that for all $s$ in the support of $S_t$ and all $z$ in the support of $S_{t-1}$,
\begin{equation}
\label{eq:bridge_eq}
\begin{aligned}
&p(A_t=a\mid S_{t-1}=z,S_t=s)\\
&=
\int h_a(w,s)\,p(w\mid S_{t-1}=z,S_t=s)\,dw.
\end{aligned}
\end{equation}
\end{assumption}
Equation~\eqref{eq:bridge_eq} is a Fredholm integral equation of the first kind. The solution links observed and counterfactual distribution. and solutions need not be unique in general.

\begin{theorem}
\label{thm:identification-continuous}
Under Assumptions
\ref{assump:proxy_ci},
\ref{assum:consistency}, 
\ref{assum:positivity},
\ref{assum:latent_exch},
\ref{assump:completeness_cont}, and \ref{assump:bridge}, the interventional action probabilities are identified by
\begin{equation}
\label{eq:identification-continuous}
p\!\left(A_t^{(s)}=a\right)
=
\int h_a(w,s)\,p(w)\,dw,
\qquad a\in\mathcal{A}.
\end{equation}
Consequently, $\pi_{\mathrm{opt}}(s)\in\arg\max_{a\in\mathcal{A}} p(A_t^{(s)}=a)$ is identified for all $s$ (up to ties).
\end{theorem}
Equipped with the identification results in
Theorems~\ref{thm:identification-discrete} and \ref{thm:identification-continuous}, the next section develops estimation procedures for $\pi_{\mathrm{opt}}$ in both continuous and discrete regimes.

\section{Estimation of the Optimal Policy}
\label{sec:est}

This section describes how to estimate the causal-optimal imitation policy
\(
\pi_{\mathrm{opt}}(s)
\)
from offline expert demonstrations.
We assume we observe \(n\) expert trajectories of length \(T\),
\(
\mathcal D_E=\{(S_{1:T}^{(i)},W_{0:T-1}^{(i)},A_{1:T}^{(i)})\}_{i=1}^n
\), and we form the pooled sample of tuples
$\mathcal I
:=\{(Z_t^{(i)},S_t^{(i)},W_{t-1}^{(i)},A_t^{(i)}): i=1,\dots,n,\ t=1,\dots,T\}$, where $Z_t=S_{t-1}$.

\subsection{Discrete Case: Plug-in Estimator with Coarsenings}

Assume \(\mathcal S,\mathcal W,\mathcal A\) are finite.
As discussed in Section~\ref{sec:id}, identification only requires \emph{some} pair of coarsenings
$Z'_t=\phi_Z(Z_t)\in\{1,\dots,m\}$,
$W'_{t-1}=\phi_W(W_{t-1})\in\{1,\dots,m\}$,
such that, for each \(s\in\mathcal S\), the matrix
\(
P_{W'\mid Z',s}\in\mathbb R^{m\times m}
\)
is invertible. Equivalently, we require \(|\mathcal Z'|=|\mathcal W'|=m\) and \(\det(P_{W'\mid Z',s})\neq 0\).
For each \(s\in\mathcal S\), estimate by empirical frequencies 
$\widehat P_{A\mid Z',s}\in\mathbb R^{|\mathcal A|\times m}$,
$\widehat P_{W'\mid Z',s}\in\mathbb R^{m\times m}$,
$\widehat p_{W'}\in\mathbb R^{m}$.
The discrete identification formula yields the natural plug-in estimator
\begin{equation}
\label{eq:plugin_discrete}
\widehat p(A_t^{(s)}) \;=\;
\widehat P_{A\mid Z',s}\;
\widehat P_{W'\mid Z',s}^{-1}\;
\widehat P_{W'} \;\in\;\mathbb R^{|\mathcal A|}.
\end{equation}
Finally, define
$\widehat \pi_{\mathrm{opt}}(s)
\;:=\;
\arg\max_{a\in\mathcal A}\;
\widehat p(A_t^{(s)}=a)$,
where \(\widehat p(A_t^{(s)}=a)\) is the \(a\)-th coordinate of \(\widehat p(A_t^{(s)})\).

\subsection{Continuous Case: RKHS-Adversarial Estimation}

Now assume states take values in continuous spaces.
For simplicity, suppose \(\mathcal A=\{1,\dots,K\}\) is finite and define the one-vs-all labels
\(
Y_t^{(a)}:=\mathbb I\{A_t=a\}
\)
for each \(a\in\mathcal A\).
The continuous identification result in Section~\ref{sec:id} implies that it suffices to estimate, for each \(a\),
a bridge function \(h_a:\mathcal W\times\mathcal S\to\mathbb R\) satisfying the conditional moment restriction
$\mathbb E\![\,Y_t^{(a)}-h_a(W_{t-1},S_t)\ |\ Z_t,S_t]=0$, a.s.
Given such \(h_a\), the causal action probability is
\begin{equation}
\label{eq:intervention_from_h}
p(A_t^{(s)}\!=\!a)
\!=\!\!\!\int \!h_a(w,s)\,p(w)\,dw
\!=\!\mathbb E\!\left[h_a(W_{t-1},s)\right],
\end{equation}
and therefore
\(
\pi_{\mathrm{opt}}(s)=\arg\max_{a\in\mathcal A}\mathbb E[h_a(W_{t-1},s)].
\)

\begin{figure*}[t] 
    \centering
    \includegraphics[width=0.3\textwidth]{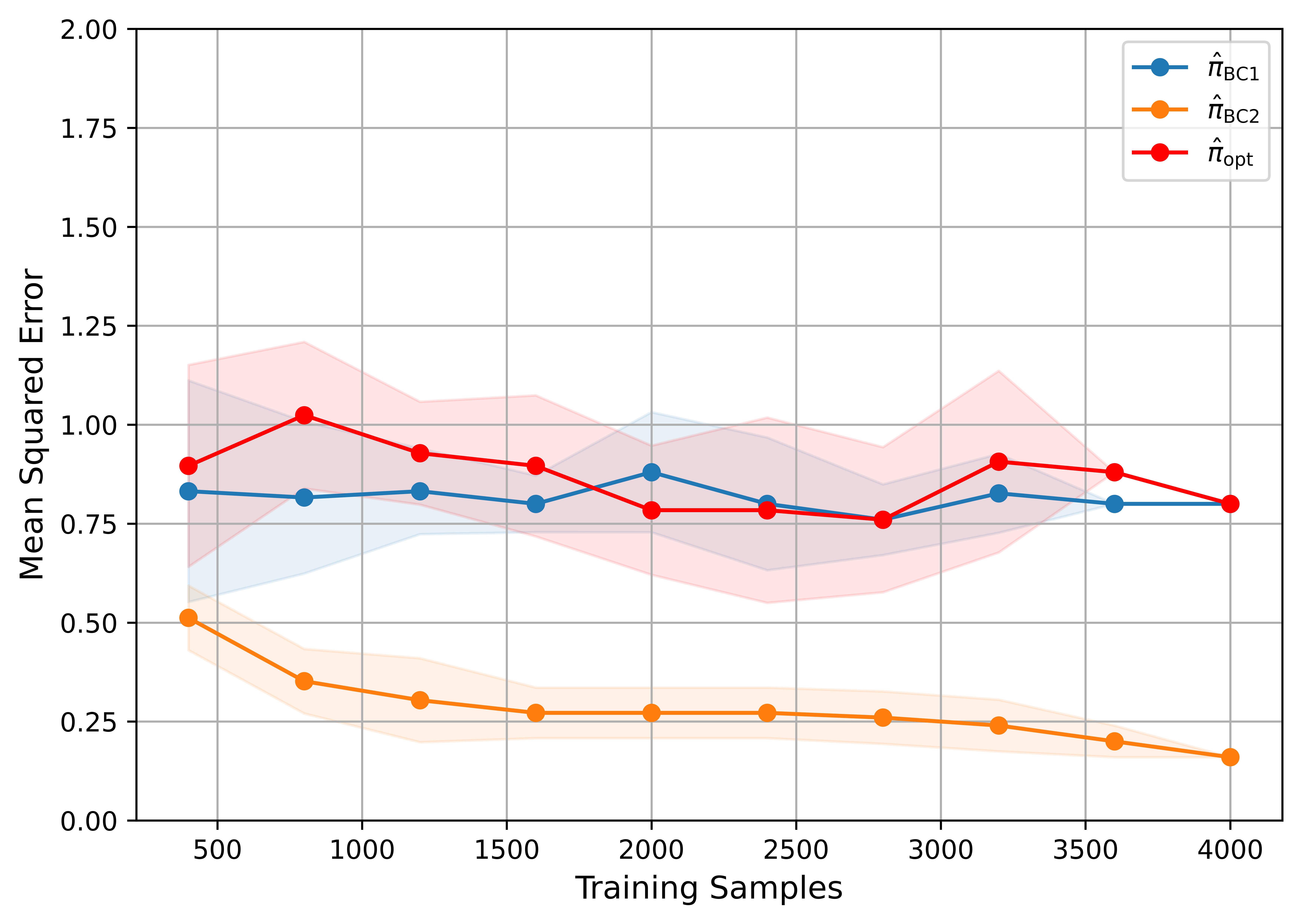}
    \hspace{.2cm}
    \includegraphics[width=0.3\textwidth]{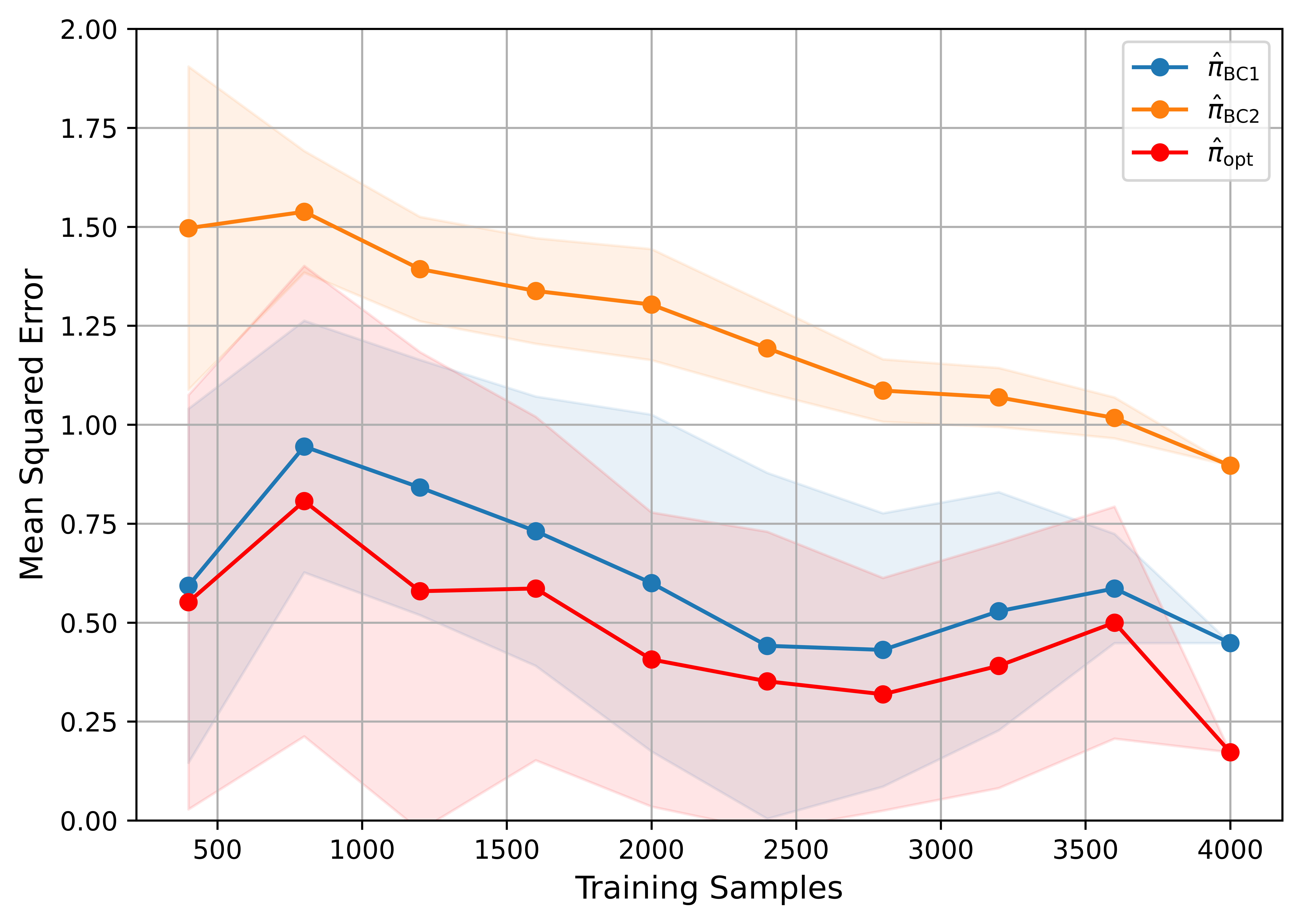}
    \hspace{.2cm}
    \includegraphics[width=0.3\textwidth]{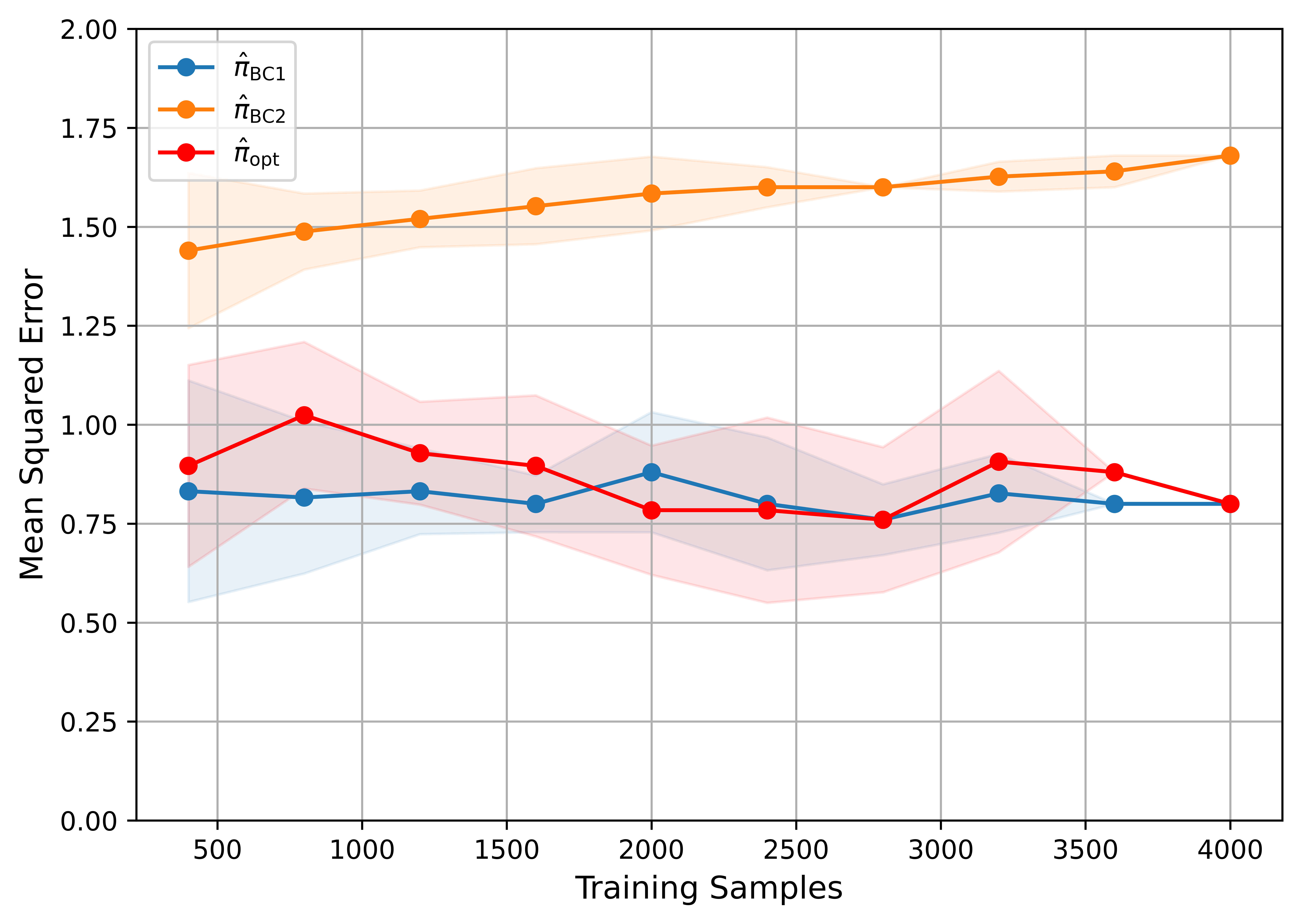}
\caption{
Mean squared error (MSE) of different policies under distributional shift.
From left to right: (i) no distributional shift; 
(ii) shift in ICU time and lactate level; 
(iii) simultaneous shifts in $P(W_{t-1}\mid U_{t-1})$.
Red corresponds to $\hat \pi_{\mathrm{opt}}$, orange to $\hat \pi_{\mathrm{BC}1}$, and green to $\hat \pi_{\mathrm{BC2}}$.
}
\label{fig:distshift_real_data}

\end{figure*}

\paragraph{Regularized minimax estimator.}
Following minimax/RKHS approach \citep{ghassami2022minimax, dikkala2020minimax} for solving integral equations in proximal causal inference,
we estimate \(h_a\) by the regularized saddle-point problem
\begin{equation}
\label{eq:minimax_specific}
\begin{aligned}
\widehat h_a
\in
&\arg\min_{h\in\mathcal H}\;
\sup_{q\in\mathcal Q}
\Big\{
\widehat{\mathbb E}\Big[
\big(Y^{(a)}-h(W,S)\big)\,q(Z,S)\\
& \;-\; q(Z,S)^2
\Big]
-\lambda_Q\|q\|_{\mathcal Q}^2
+\lambda_H\|h\|_{\mathcal H}^2
\Big\},
\end{aligned}
\end{equation}
where \(\widehat{\mathbb E}\) denotes the empirical average over the pooled tuples in \(\mathcal I\),
\(\mathcal H\) is an RKHS over \(\mathcal W\times\mathcal S\) (for \(h\)),
\(\mathcal Q\) is an RKHS over \(\mathcal Z\times\mathcal S\) (for \(q\)),
and \(\lambda_Q,\lambda_H>0\) are regularization parameters.
The regularization parameters
control the smoothness of the estimators that can be tuned via cross validation on projected error (see \citep{ghassami2022minimax, dikkala2020minimax}).

Let \(N:=|\mathcal I|=nT\) be the number of pooled tuples and index them by \(j=1,\dots,N\),
writing \((Z_j,S_j,W_j,A_j)\) and \(Y_j^{(a)}=\mathbb I\{A_j=a\}\).
Let \(K_H\in\mathbb R^{N\times N}\) and \(K_Q\in\mathbb R^{N\times N}\) be the empirical kernel matrices
$(K_H)_{ij}=K_{\mathcal H}\big((W_i,S_i),(W_j,S_j)\big)$,
$(K_Q)_{ij}=K_{\mathcal Q}\big((Z_i,S_i),(Z_j,S_j)\big)$,
and define
$\Gamma
:=\frac14\,K_Q\left(\frac1N K_Q+\lambda_Q I_N\right)^{-1}$.

\begin{proposition}[Closed-form solution]
\label{prop:rkhs_closed_form_causil}
Assume \(\mathcal H\) and \(\mathcal Q\) are RKHSs and consider \eqref{eq:minimax_specific}.
Then \(\widehat h_a\) admits a representer 
\[
\widehat h_a(w,s)=\sum_{j=1}^N \alpha_{a,j}\,
K_{\mathcal H}\big((W_j,S_j),(w,s)\big)
\]
for coefficient vector \(\alpha_a\in\mathbb R^N\) defined as
\begin{equation}
\label{eq:alpha_closed_form}
\alpha_a
=
\left(K_H\Gamma K_H + N^2\lambda_H K_H\right)^{\dagger}\,K_H\Gamma\,y_a,
\end{equation}
where \(y_a:=(Y^{(a)}_1,\dots,Y^{(a)}_N)^\top\) and \((\cdot)^{\dagger}\) denotes the Moore--Penrose pseudoinverse.
\end{proposition}
Given \(\widehat h_a\), we estimate \eqref{eq:intervention_from_h} by the empirical plug-in
$\widehat p(A_t^{(s)}=a)
\;:=\;
\frac1N\sum_{j=1}^N \widehat h_a(W_j,s)$, $a\in\mathcal A$,
and return
$\widehat\pi_{\mathrm{opt}}(s)
\;:=\;
\arg\max_{a\in\mathcal A}\;\widehat p(A_t^{(s)}=a)$.

A pseudocode covering both the discrete and continuous estimators is presented in Algorithm \ref{alg:causil}.

\begin{algorithm}[t]
\caption{\texttt{CausIL}: Estimating $\pi_{\mathrm{opt}}$ (discrete or continuous)}
\label{alg:causil}
\begin{algorithmic}[1]
\REQUIRE Expert trajectories $\mathcal D_E=\{(S_{1:T}^{(i)},W_{0:T-1}^{(i)},A_{1:T}^{(i)})\}_{i=1}^n$; mode $\in\{\textsc{Discrete},\textsc{Continuous}\}$.
\ENSURE Estimated policy $\widehat\pi_{\mathrm{opt}}(\cdot)$.

\STATE Form pooled tuples $\mathcal I=\{(Z_t^{(i)},S_t^{(i)},W_{t-1}^{(i)},A_t^{(i)})\}$ with $Z_t:=S_{t-1}$; let $N:=|\mathcal I|$.

\IF{mode $=$ \textsc{Discrete}}
    \STATE Choose coarsenings $Z'_t=\phi_Z(Z_t)$ and $W'_{t-1}=\phi_W(W_{t-1})$ such that $\widehat P_{W'\mid Z',s}$ is invertible for each $s\in\mathcal S$.
    \FOR{each $s\in\mathcal S$}
        \STATE Estimate $\widehat P_{A\mid Z',s}$, $\widehat P_{W'\mid Z',s}$, and $\widehat p_{W'}$ by empirical frequencies over $\mathcal I$.
        \STATE Compute $\widehat p(A^{(s)}) \leftarrow \widehat P_{A\mid Z',s}\;\widehat P_{W'\mid Z',s}^{-1}\;\widehat P_{W'}$. (Eq.~\eqref{eq:plugin_discrete})
        \STATE Set $\widehat\pi_{\mathrm{opt}}(s)\leftarrow \arg\max_{a\in\mathcal A}\widehat p(A^{(s)}=a)$.
    \ENDFOR
\ELSE
    \STATE Fix RKHSs $\mathcal H$ on $\mathcal W\times\mathcal S$ and $\mathcal Q$ on $\mathcal Z\times\mathcal S$, with kernels $K_{\mathcal H},K_{\mathcal Q}$ and regularization $\lambda_H,\lambda_Q>0$.
    \STATE Build kernel matrices $K_H,K_Q\in\mathbb R^{N\times N}$ over $\mathcal I$ and compute $\Gamma\leftarrow \frac14 K_Q\left(\frac1N K_Q+\lambda_Q I_N\right)^{-1}$.
    \FOR{each action $a\in\mathcal A$}
        \STATE Set $y_a[j]\leftarrow \mathbb I\{A_j=a\}$ for pooled index $j=1,\dots,N$.
        \STATE Compute $\alpha_a \leftarrow (K_H\Gamma K_H + N^2\lambda_H K_H)^{\dagger}\,K_H\Gamma\,y_a$. (Eq.~\eqref{eq:alpha_closed_form})
        \STATE Define $\widehat h_a(w,s)\leftarrow \sum_{j=1}^N \alpha_{a,j}\,K_{\mathcal H}((W_j,S_j),(w,s))$. (Proposition~\ref{prop:rkhs_closed_form_causil})
    \ENDFOR
    \STATE For a query $s$, compute $\widehat p(A^{(s)}=a)\leftarrow \frac1N\sum_{j=1}^N \widehat h_a(W_j,s)$ for all $a\in\mathcal A$. 
    \STATE Set $\widehat\pi_{\mathrm{opt}}(s)\leftarrow \arg\max_{a\in\mathcal A}\widehat p(A^{(s)}=a)$.
\ENDIF

\end{algorithmic}
\end{algorithm}

\section{Experiments}
\label{sec:experiments}

\subsection{Simulation study}

We generate $n$ expert trajectories of length $T$ from the model in Section~\ref{sec:model}.
In simulation, the full data include latent variables,
\(
\{(S_{1:T}^{(i)},W_{0:T-1}^{(i)},U_{0:T-1}^{(i)},A_{1:T}^{(i)})\}_{i=1}^n,
\)
but the learner observes only \((S_t,W_{t-1},A_t)\).
We pool decision-time tuples
\(
\mathcal I:=\{(Z_t^{(i)},S_t^{(i)},W_{t-1}^{(i)},A_t^{(i)}): i=1,\dots,n,\ t=1,\dots,T\},
\)
with $Z_t:=S_{t-1}$,
and all variables are categorical (with possibly different cardinalities).
Briefly, $W_{t-1}$ is generated from $U_{t-1}$, the latent state evolves as a function of $(U_{t-1},S_{t-1},A_{t-1})$, the observed state depends on $(U_t,U_{t-1},S_{t-1},A_{t-1})$, and the expert acts according to $A_t=\pi_E(S_t,U_{t-1})$.
Exact functional forms and parameters are reported in the appendix.

We vary the training size from $N=100$ to $N=1000$ and evaluate on a fixed test set of $N_{\mathrm{test}}=1000$.
As mentioned in Section~\ref{sec:model}, we consider two types of test-time shift: (i) \emph{measurement shift} in the proxy channel $p(W_{t-1}\mid U_{t-1})$, and (ii) \emph{dynamics shift} in the state/latent transition mechanisms.
Performance is measured by the mean squared one-hot imitation loss
\[
\mathrm{MSE}(\hat\pi)
=
\frac{1}{N_{\mathrm{test}}}
\sum_{j=1}^{N_{\mathrm{test}}}
\big\|e(A_j)-e(\hat\pi(X_j))\big\|_2^2,
\]
where $(X_j,A_j)$ are test tuples (with $X_j=S_t$ for BC1/\texttt{CausIL} and $X_j=(S_t,V_{t-1})$ for BC2), and $e(\cdot)$ is the one-hot encoding.
Fig.~\ref{fig:distshift} summarizes results: \texttt{CausIL} is robust to measurement shift and, under dynamics shifts that preserve the marginal prevalence of the delayed latent context, is empirically more stable than BC baselines.

\subsection{Real-Data Experiment}

We evaluate on the PhysioNet/Computing in Cardiology Challenge 2019 cohort \citep{reyna2020early}, which provides hourly ICU time series of vital signs and laboratory measurements from multiple hospitals.
Because the released data do not include explicit clinician intervention actions (e.g., fluids or vasopressors), we construct a \emph{semi-simulated} imitation task using real covariate trajectories and simulated expert actions.

We define the observed state using a set of routinely monitored vitals,
\(
S_t=(\texttt{HR}_t,\texttt{MAP}_t,\texttt{O2Sat}_t,\texttt{Resp}_t,\texttt{Temp}_t),
\)
and treat lactate as a delayed latent context $U_t:=\texttt{Lactate}_t$ that reflects underlying perfusion/severity but is typically incorporated with delay (e.g., due to lab turnaround and charting).
We simulate a three-level ``hemodynamic support intensity'' action
\(A_t\in\{0,1,2\}\) (no escalation / moderate escalation / aggressive escalation)
via an expert rule $A_t=\pi_E(S_t,U_{t-1})$, and provide the learner a noisy proxy $W_{t-1}$ of $U_{t-1}$.
To study robustness, we induce (i) \emph{population shift} by constructing a test cohort with systematically different trajectory composition (e.g., longer stays and higher lactate prevalence), and (ii) \emph{measurement shift} by changing the noise/bias in the measurement mechanism $p(W_{t-1}\mid U_{t-1})$ between train and test.
Full preprocessing and simulation details are in Appendix~\ref{appendix:data_preprocessing}.

Fig.~\ref{fig:distshift_real_data} reports imitation error under both shift types.
BC baselines exhibit increased error and variability under population shift, and BC2 degrades sharply under measurement shift due to reliance on the measurement mechanism.
In contrast, \texttt{CausIL} attains lower error and remains stable across training sizes.

\section{Conclusion}

We studied offline imitation learning under state measurement error and distribution shift, and showed that BC can converge to systematically biased policies because it relies on non-invariant observational state--action correlations. 
To address this, we proposed a causal imitation framework establishing identification conditions for a causal-optimal target policy and developing estimators for both discrete and continuous settings.
Across simulation studies and a semi-simulated ICU decision-making task built from the PhysioNet/Computing in Cardiology Challenge 2019 cohort, \texttt{CausIL} consistently achieved lower imitation error and exhibited markedly improved stability under measurement-mechanism and population shifts relative to BC baselines. 
These results highlight the importance of targeting interventional quantities when learning policies from purely observational demonstrations in realistic, shifting environments.

\newpage
\bibliographystyle{unsrtnat}
\bibliography{references}

\appendix
\onecolumn

\begin{center}
~\\
{\bf\Large Appendix for}

{\bf\Large ``Causal Imitation Learning Under Measurement Error and Distribution Shift''}
\end{center}

\section{More on Related Work}
In this section, we provide a more detailed discussion of related work to complement the concise overview in the main text. 
We elaborate on prior literature in imitation learning, off-policy evaluation under partial observability, and causal imitation learning.

\textbf{Imitation Learning.}
Imitation learning (IL) studies the problem of learning a policy from expert demonstrations~\citep{pomerleau1988alvinn, lecun2006tutorial, russell1998learning, ho2016generative}. 
Broadly, IL methods can be categorized into \emph{offline}, \emph{online}, and \emph{interactive} approaches. 
Offline methods such as Behavioral Cloning (BC)~\citep{pomerleau1988alvinn} learn directly from collected expert trajectories without interaction, while inverse reinforcement learning~\citep{russell1998learning} and adversarial IL methods~\citep{ho2016generative} attempt to infer the expert’s underlying reward or occupancy distribution. 
Interactive IL algorithms, such as DAgger~\citep{ross2011reduction} and AggreVaTe~\citep{ross2014reinforcement}, extend standard BC by querying an interactive expert to correct errors accumulated during learning. 
However, such interaction is often infeasible in real-world settings where only a fixed set of demonstrations is available. 
Our work therefore focuses on the offline setting, but differs from standard IL methods by addressing the presence of unobserved confounders and measurement error in state observations. 
We aim to recover what the expert \emph{intended} to do in each true state, rather than imitating the corrupted actions observed under noisy measurements.

\textbf{Off-policy Evaluation in POMDPs.}
A parallel line of work studies off-policy evaluation in partially observable or confounded environments, typically modeled as Partially Observable Markov Decision Processes (POMDPs) with latent states. 
\cite{tennenholtz2020off} introduce a decoupled POMDP formulation in which an additional observation channel enables recovery of the latent state distribution, allowing unbiased value estimation under strong invertibility assumptions. 
\cite{miao2022off} study episodic POMDPs under nonparametric models and show that the value of a target policy can be identified using two proxy variables associated with the hidden state through bridge-function-based arguments. 
Subsequent work extends off-policy evaluation to more general confounded POMDPs and Markov Decision Processes, including minimax-based value estimation~\citep{shi2022minimax}, front-door identification and confidence interval construction~\citep{shi2024off}, proximal learning for partially observed systems~\citep{bennett2024proximal}, and model-based approaches that deconfound both rewards and transitions~\citep{hong2024model}. 
Across these settings, the primary goal is to estimate or evaluate policy values using reward signals, and confounding is modeled as contemporaneous, with latent variables affecting actions and rewards at the same time step. 
In contrast, our work focuses on imitation learning without rewards, where actions have delayed effects on latent variables and proxy information arises from measurement error in state observations. 
This leads to a different causal structure and shifts the focus from value estimation to recovering the expert policy.

\textbf{Imitation Learning with Latent Confounding.}
Recent work has increasingly framed imitation learning as a causal inference problem, emphasizing that expert demonstrations may be biased by hidden variables that jointly influence states and actions. 
Early studies revealed that such latent factors induce \emph{causal delusion}, where sequence models incorrectly treat their own actions as evidence about unobserved context~\citep{ortega2021shaking}. 
This insight motivated a line of interventional IL methods~\citep{vuorio2022deconfounded, swamy2022sequence}, which correct posterior updates by treating past actions as interventions and rely on on-policy expert feedback such as DAgger~\citep{ross2011reduction}. 

Orthogonal to these interactive approaches, several works assume that expert demonstrations are corrupted by latent confounders (possibly unknown even to the expert) and formulate imitation learning as an instrumental-variable problem~\citep{swamy2021moments, shao2025unifying, zeng2025confounded}. 
By mitigating confounding, often due to temporally correlated noise, these methods enable recovery of the underlying policy without requiring an interactive expert. 
Other efforts study theoretical identifiability of value-equivalent imitation policies using back-door or front-door adjustments on causal graphs~\citep{zhang2020causal, kumor2021sequential}. 
A complementary literature examines \emph{causal confusion} arising from spurious correlations in the observation space despite the absence of latent variables~\citep{de2019causal, pfrommer2023initial, wen2020fighting, spencer2021feedback}. 

As summarized in Table~\ref{tab:comparison1}, prior causal IL methods typically rely on graphical adjustment, interactive expert correction, or instrumental-variable formulations designed for short-range confounding. 
These approaches do not account for measurement error in state observations, which induces a genuinely temporal confounding structure with delayed effects on actions. 
In contrast, we reinterpret noisy measurements as proxy variables and derive proximal conditional moment restrictions that recover the expert’s causal policy without interventions or valid instruments, enabling fully offline imitation beyond the scope of existing causal IL frameworks.

\section{A Simple DGP Illustrating $\pi_{\mathrm{opt}}$ vs.\ Behavioral Cloning}
\label{app:dgp_bc_vs_opt}

We provide a simple population example showing that $\pi_{\mathrm{BC1}}$, $\pi_{\mathrm{BC2}}$, and $\pi_{\mathrm{opt}}$ can disagree even with infinite data.
Consider scalar latent context $U_{t-1}\in\mathbb{R}$, observed state $S_t\in\mathbb{R}$, proxy measurement $W_{t-1}\in\mathbb{R}$, and binary actions $\mathcal{A}=\{0,1\}$.
Let
\begin{align*}
U_{t-1} &\sim \mathcal{N}(\mu,\sigma_U^2),\\
S_t &= \alpha\,U_{t-1}+\varepsilon_S,\qquad \varepsilon_S\sim \mathcal{N}(0,\sigma_S^2),\\
W_{t-1} &= U_{t-1}+\varepsilon_W,\qquad \varepsilon_W\sim \mathcal{N}(0,\sigma_W^2),
\end{align*}
with $(\varepsilon_S,\varepsilon_W)\indep U_{t-1}$.
The expert uses both the current observed state and the delayed latent context via a threshold rule
\begin{equation}
A_t=\pi_E(S_t,U_{t-1})
=\mathbb{I}\!\left\{\beta_S S_t+\beta_U U_{t-1}\ge c\right\},
\label{eq:app_expert_rule}
\end{equation}
for fixed coefficients $\beta_S,\beta_U$ and threshold $c$, and assume $\beta_U>0$.

\paragraph{Causal target $\pi_{\mathrm{opt}}$.}
Under the intervention $\mathrm{do}(S_t=s)$, we have
$A_t^{(s)}=\mathbb{I}\{\beta_S s+\beta_U U_{t-1}\ge c\}$.
Therefore,
\[
\mathbb{P}(A_t^{(s)}=1)
=\mathbb{P}\!\left(U_{t-1}\ge \frac{c-\beta_S s}{\beta_U}\right).
\]
Since $\mathcal{A}=\{0,1\}$, the interventional argmax satisfies
$\pi_{\mathrm{opt}}(s)=1$ iff $\mathbb{P}(A_t^{(s)}=1)\ge \tfrac12$, i.e., iff the threshold is below the median of $U_{t-1}$.
For a Gaussian, the median equals the mean, so
\begin{equation}
\pi_{\mathrm{opt}}(s)
=\mathbb{I}\!\left\{\beta_S s+\beta_U \mu \ge c\right\}.
\label{eq:app_piopt}
\end{equation}

\paragraph{BC1 target $\pi_{\mathrm{BC1}}$.}
BC1 is defined by $\pi_{\mathrm{BC1}}(s)=1$ iff $\mathbb{P}(A_t=1\mid S_t=s)\ge \tfrac12$.
Because $(U_{t-1},S_t)$ are jointly Gaussian, $U_{t-1}\mid S_t=s$ is Gaussian with mean
\begin{equation}
m_1(s):=\mathbb{E}[U_{t-1}\mid S_t=s]
=\mu+\frac{\alpha\sigma_U^2}{\alpha^2\sigma_U^2+\sigma_S^2}\,\big(s-\alpha\mu\big).
\label{eq:app_m1}
\end{equation}
Moreover,
\begin{align*}
&\mathbb{P}(A_t\!=\!1\mid S_t\!=\!s)
\!=\!\mathbb{P}\!\left(\beta_S s+\beta_U U_{t-1}\ge c\mid S_t\!=\!s\right)
\ge \tfrac12\\
&\iff
\beta_S s+\beta_U m_1(s)\ge c,
\end{align*}
so
\begin{equation}
\pi_{\mathrm{BC1}}(s)
=\mathbb{I}\!\left\{\beta_S s+\beta_U m_1(s)\ge c\right\}.
\label{eq:app_pibc1}
\end{equation}
Comparing \eqref{eq:app_piopt} and \eqref{eq:app_pibc1}, whenever $\alpha\neq 0$ and $\beta_U\neq 0$, the posterior mean $m_1(s)$ depends on $s$, so the decision boundary of $\pi_{\mathrm{BC1}}$ generally differs from that of $\pi_{\mathrm{opt}}$.

\paragraph{BC2 target $\pi_{\mathrm{BC2}}$ and sensitivity to measurement mechanism.}
BC2 conditions on both $S_t$ and the proxy $W_{t-1}$.
Define $V_{t-1}=(S_{t-1},W_{t-1})$; in this static illustration we focus on the dependence on $W_{t-1}$ and write $\pi_{\mathrm{BC2}}(s,w)$.
Again by joint Gaussianity, $U_{t-1}\mid (S_t=s,W_{t-1}=w)$ is Gaussian with mean
\begin{equation}
\begin{aligned}
m_2(s,w):=\mathbb{E}[U_{t-1}\mid S_t=s, W_{t-1}=w]
=\mu + \Sigma_{U,Y}\Sigma_{Y,Y}^{-1}\!\left(\begin{bmatrix}s\\w\end{bmatrix}-\begin{bmatrix}\alpha\mu\\\mu\end{bmatrix}\right),
\end{aligned}
\label{eq:app_m2_general}
\end{equation}
where $Y=(S_t,W_{t-1})$,
$\Sigma_{U,Y}=(\alpha\sigma_U^2,\ \sigma_U^2)$, and
\[
\Sigma_{Y,Y}=
\begin{bmatrix}
\alpha^2\sigma_U^2+\sigma_S^2 & \alpha\sigma_U^2\\
\alpha\sigma_U^2 & \sigma_U^2+\sigma_W^2
\end{bmatrix}.
\]
As before, $\pi_{\mathrm{BC2}}(s,w)=1$ iff $\beta_S s+\beta_U m_2(s,w)\ge c$, hence
\begin{equation}
\pi_{\mathrm{BC2}}(s,w)
=\mathbb{I}\!\left\{\beta_S s+\beta_U m_2(s,w)\ge c\right\},
\label{eq:app_pibc2}
\end{equation}
which generally depends on $w$ and therefore cannot coincide with the state-only target $\pi_{\mathrm{opt}}(s)$.
Furthermore, the conditional mean $m_2(s,w)$ depends on the measurement noise variance $\sigma_W^2$ through $\Sigma_{Y,Y}^{-1}$; thus changes in the measurement mechanism (e.g., different devices across domains, modeled as different $\sigma_W^2$) can change $\pi_{\mathrm{BC2}}$ even when the expert rule \eqref{eq:app_expert_rule} and the causal target \eqref{eq:app_piopt} remain unchanged.

\paragraph{Connection to Proposition~\ref{prop:rare_coincidence}.}
In the DGP above, when $\alpha\neq 0$ we have $S_t=\alpha U_{t-1}+\varepsilon_S$, so $U_{t-1}\not\!\perp\!\!\!\perp S_t$ and therefore the sufficient condition for $\pi_{\mathrm{BC1}}=\pi_{\mathrm{opt}}$ in Proposition~\ref{prop:rare_coincidence}(1) fails; correspondingly, \eqref{eq:app_piopt} and \eqref{eq:app_pibc1} yield different decision boundaries unless $\beta_U=0$ (a degenerate case where the expert ignores $U_{t-1}$).
Likewise, since $W_{t-1}=U_{t-1}+\varepsilon_W$ carries information about $U_{t-1}$, we also have $U_{t-1}\not\!\perp\!\!\!\perp (S_t,W_{t-1})$ for non-degenerate noise, so the sufficient condition for $\pi_{\mathrm{BC2}}(s,v)=\pi_{\mathrm{opt}}(s)$ in Proposition~\ref{prop:rare_coincidence}(2) fails as well.
This illustrates concretely that observational BC targets coincide with the causal target only in rare settings where conditioning on observed covariates does not change the latent context distribution (or where the expert policy effectively does not depend on the latent context).

\section{Proofs}
 \label{Appendix}

\subsection{Proof of Proposition \ref{prop:bc_l2_det}}
For an arbitrary deterministic policy $\pi:\mathcal X\to\mathcal A$.
Since $A_t\in\mathcal A=\{1,\dots,K\}$ is categorical and $e(\cdot)$ denotes the
one-hot encoding, we have
\[
\|e(A_t)-e(\pi(X_t))\|_2^2
=
\begin{cases}
0, & A_t=\pi(X_t),\\
2, & A_t\neq \pi(X_t),
\end{cases}
\]
because two distinct one-hot vectors in $\mathbb R^K$ differ in exactly two
coordinates.

Taking expectation under $(X_t,A_t)\sim d_{\pi_E}$ yields
\begin{align*}
\mathcal R(\pi)
&=\mathbb E\!\left[\|e(A_t)-e(\pi(X_t))\|_2^2\right] \\
& = 2\,\mathbb E[\mathbf 1\{A_t\neq \pi(X_t)\}]\\
&= 2\,\mathbb P\!\left(A_t\neq \pi(X_t)\right).
\end{align*}
Using the law of total expectation, this can be written as
\begin{align*}
\mathcal R(\pi)
&= 2\,\mathbb E\!\left[
\mathbb P(A_t\neq \pi(X_t)\mid X_t)
\right] \\
&= 2\,\mathbb E\!\left[
1 - p(A_t=\pi(X_t)\mid X_t)
\right].
\end{align*}
Since the outer expectation is taken with respect to the marginal distribution of $X_t$, minimizing $\mathcal R(\pi)$ over all deterministic policies $\pi$
is equivalent to minimizing the 
\(
1 - p(A_t=\pi(x)\mid X_t=x)
\)
for almost every $x\in\mathcal X$.
Equivalently, for almost every $x$, any risk minimizer $\pi^\star$ must satisfy
\[
\pi^\star(x)\in
\arg\max_{a\in\mathcal A} p(A_t=a\mid X_t=x).
\]
Therefore, any $\pi^\star\in\arg\min_\pi \mathcal R(\pi)$ coincides almost
everywhere with a Bayes classifier that selects an action maximizing the conditional action probability.

\subsection{Proof of Proposition \ref{prop:piopt_core}}
Fix any $s\in\mathcal S$ and $a\in\mathcal A$.  
By definition, the causal imitation target is
\[
\pi_{\mathrm{opt}}(s)\in\arg\max_{a\in\mathcal A}\mathbb P(A_t^{(s)}=a).
\]
Under the assumed deterministic expert mechanism $A_t=\pi_E(S_t,U_{t-1})$ and standard consistency and temporal ordering, intervening on $S_t$ at level $s$ leaves the latent variable $U_{t-1}$ unchanged and yields
\[
A_t^{(s)}=\pi_E(s,U_{t-1}).
\]
Therefore,
\[
\mathbb P(A_t^{(s)}=a)
=
\mathbb P\!\left(\pi_E(s,U_{t-1})=a\right),
\]
where the probability is taken with respect to the marginal (population) distribution of $U_{t-1}$.  
Since $\mathcal A$ is finite, the event $\{\pi_E(s,U_{t-1})=a\}$ is measurable and its probability can be written as an expectation of the corresponding indicator function, namely
\[
\mathbb P\!\left(\pi_E(s,U_{t-1})=a\right)
=
\mathbb E\!\left[\mathbb I\{\pi_E(s,U_{t-1})=a\}\right].
\]
Substituting this representation into the definition of $\pi_{\mathrm{opt}}(s)$ gives
\[
\pi_{\mathrm{opt}}(s)\in\arg\max_{a\in\mathcal A}
\mathbb E\!\left[\mathbb I\{\pi_E(s,U_{t-1})=a\}\right],
\]
which proves the claim.

\subsection{Proof of Corollary \ref{cor:pa}}
By definition,
\[
\bar{\pi}_E(a\mid s)
=
\mathbb P\!\left(\pi_E(s,U_{t-1})=a\right)
=
\mathbb E\!\left[\mathbb I\{\pi_E(s,U_{t-1})=a\}\right].
\]
Proposition~\ref{prop:piopt_core} states that
\[
\pi_{\mathrm{opt}}(s)\in\arg\max_{a\in\mathcal A}
\mathbb E\!\left[\mathbb I\{\pi_E(s,U_{t-1})=a\}\right],
\]
and therefore
\[
\pi_{\mathrm{opt}}(s)\in\arg\max_{a\in\mathcal A}\bar{\pi}_E(a\mid s).
\]
Moreover, for any fixed $a\in\mathcal A$,
\[
\mathbb E\!\left[\mathbb I\{\pi_E(s,U_{t-1})\neq a\}\right]
=
1-\mathbb E\!\left[\mathbb I\{\pi_E(s,U_{t-1})=a\}\right]
=
1-\bar{\pi}_E(a\mid s).
\]
Since subtracting from a constant does not change the optimizer, minimizing
$\mathbb E[\mathbb I\{\pi_E(s,U_{t-1})\neq a\}]$ over $a\in\mathcal A$ is equivalent to maximizing
$\bar{\pi}_E(a\mid s)$.
Consequently,
\[
\pi_{\mathrm{opt}}(s)\in\arg\min_{a\in\mathcal A}
\mathbb E\!\left[\mathbb I\{\pi_E(s,U_{t-1})\neq a\}\right],
\]
which completes the proof.

\subsection{Proof of Proposition \ref{prop:rare_coincidence}}

We prove the claim by explicitly characterizing the three policies
$\pi_{\mathrm{opt}}$, $\pi_{\mathrm{BC1}}$, and $\pi_{\mathrm{BC2}}$
in terms of probabilities of the action-induced latent sets
$B_a(s)$.
The argument proceeds as follows.
First, we show that under a deterministic expert mechanism,
each action probability can be rewritten as the probability that the latent variable
$U_{t-1}$ falls into the corresponding set $B_a(s)$.
Next, we use these identities to express the BC targets and $\pi_{\mathrm{opt}}$
as maximizers of different probability functionals over $a\in\mathcal A$.
Finally, we verify that under the stated independence conditions,
these maximization problems coincide, yielding equality of the induced policies
up to arbitrary tie-breaking.

Fix any $s\in\mathcal S$.
For each $a\in\mathcal A$, recall the action-induced latent set
\[
B_a(s):=\{u\in\mathcal U:\ \pi_E(s,u)=a\}.
\]
Since $\pi_E$ is deterministic and $\mathcal A$ is finite,
the collection $\{B_a(s)\}_{a\in\mathcal A}$ forms a partition of $\mathcal U$.

We first rewrite action probabilities in terms of the sets $B_a(s)$.
Because the expert mechanism is deterministic, for any realization,
\[
\mathbf 1\{A_t=a\}=\mathbf 1\{\pi_E(S_t,U_{t-1})=a\}.
\]
Conditioning on $S_t=s$ therefore yields
\begin{align*}
\left[\mathbf 1\{\pi_E(S_t,U_{t-1})=a\}\mid S_t=s\right]
&=\mathbf 1\{\pi_E(s,U_{t-1})=a\}
= \mathbf 1\{U_{t-1}\in B_a(s)\}.
\end{align*}
Taking conditional expectations gives
\begin{align*}
\mathbb P(A_t=a\mid S_t=s)
&=\mathbb E\big[\mathbf 1\{A_t=a\}\mid S_t=s\big] \\
&=\mathbb E\big[\mathbf 1\{\pi_E(s,U_{t-1})=a\}\mid S_t=s\big] \\
&=\mathbb E\big[\mathbf 1\{U_{t-1}\in B_a(s)\}\mid S_t=s\big] \\
&=\mathbb P(U_{t-1}\in B_a(s)\mid S_t=s).
\end{align*}
An identical argument, now conditioning on $(S_t,V_{t-1})=(s,v)$, yields
\begin{align*}
\mathbb P(A_t=a\mid S_t=s,V_{t-1}=v)
&=\mathbb E\big[\mathbf 1\{A_t=a\}\mid S_t=s,V_{t-1}=v\big] \\
&=\mathbb E\big[\mathbf 1\{\pi_E(s,U_{t-1})=a\}\mid S_t=s,V_{t-1}=v\big] \\
&=\mathbb E\big[\mathbf 1\{U_{t-1}\in B_a(s)\}\mid S_t=s,V_{t-1}=v\big] \\
&=\mathbb P(U_{t-1}\in B_a(s)\mid S_t=s,V_{t-1}=v).
\end{align*}
Without conditioning, we similarly have
\begin{align*}
\mathbb P(\pi_E(s,U_{t-1})=a)
&=\mathbb E\big[\mathbf 1\{\pi_E(s,U_{t-1})=a\}\big] \\
&=\mathbb E\big[\mathbf 1\{U_{t-1}\in B_a(s)\}\big]\\
&=\mathbb P(U_{t-1}\in B_a(s)).
\end{align*}

We now characterize the three policies via maximization over $a\in\mathcal A$.
By definition, the BC1 target satisfies
\[
\pi_{\mathrm{BC1}}(s)\in\arg\max_{a\in\mathcal A}\mathbb P(A_t=a\mid S_t=s),
\]
where $\arg\max$ denotes the (nonempty) set of maximizers and a deterministic
policy is obtained by applying an arbitrary tie-breaking rule when this set
contains more than one element.
Substituting the identity above, we obtain
\[
\pi_{\mathrm{BC1}}(s)\in
\arg\max_{a\in\mathcal A}
\mathbb P(U_{t-1}\in B_a(s)\mid S_t=s),
\]
up to ties.
Likewise, by definition of the BC2 target,
\[
\pi_{\mathrm{BC2}}(s,v)\in\arg\max_{a\in\mathcal A}
\mathbb P(A_t=a\mid S_t=s,V_{t-1}=v),
\]
which is equivalent to
\[
\pi_{\mathrm{BC2}}(s,v)\in
\arg\max_{a\in\mathcal A}
\mathbb P(U_{t-1}\in B_a(s)\mid S_t=s,V_{t-1}=v),
\]
again up to arbitrary tie-breaking.

Next, we characterize $\pi_{\mathrm{opt}}$.
Under the intervention $\mathrm{do}(S_t=s)$, the latent state $U_{t-1}$ is not
affected, and the expert continues to act according to the same deterministic
mechanism.
Thus,
\[
A_t^{(s)}=\pi_E(s,U_{t-1}),
\]
and for any $a\in\mathcal A$,
\begin{align*}
\mathbb P(A_t^{(s)}=a)
&=\mathbb P(\pi_E(s,U_{t-1})=a)
=\mathbb P(U_{t-1}\in B_a(s)).
\end{align*}
By definition,
\[
\pi_{\mathrm{opt}}(s)\in\arg\max_{a\in\mathcal A}\mathbb P(A_t^{(s)}=a),
\]
which is therefore equivalent to
\[
\pi_{\mathrm{opt}}(s)\in
\arg\max_{a\in\mathcal A}
\mathbb P(U_{t-1}\in B_a(s)),
\]
up to ties.

Finally, we verify the sufficient conditions for coincidence.
If $U_{t-1}\perp\!\!\!\perp S_t$, then for every $a\in\mathcal A$,
\[
\mathbb P(U_{t-1}\in B_a(s)\mid S_t=s)
=\mathbb P(U_{t-1}\in B_a(s)),
\]
so the objective functions inside the $\arg\max$ definitions of
$\pi_{\mathrm{BC1}}(s)$ and $\pi_{\mathrm{opt}}(s)$ coincide pointwise in $a$.
Hence the sets of maximizers are identical, and after tie-breaking,
$\pi_{\mathrm{BC1}}(s)=\pi_{\mathrm{opt}}(s)$ for all $s$.

Similarly, if $U_{t-1}\perp\!\!\!\perp (S_t,V_{t-1})$, then for every $(s,v)$ and
every $a\in\mathcal A$,
\[
\mathbb P(U_{t-1}\in B_a(s)\mid S_t=s,V_{t-1}=v)
=\mathbb P(U_{t-1}\in B_a(s)),
\]
which implies equality of the corresponding $\arg\max$ sets and thus,
up to ties,
\[
\pi_{\mathrm{BC2}}(s,v)=\pi_{\mathrm{opt}}(s)
\qquad \text{for all }(s,v).
\]
This completes the proof.

\subsection{Proof of Theorem \ref{thm:identification-discrete}}

By Assumption~\ref{assump:proxy_ci}, we can write
\begin{align}
P_{A\mid Z,s}
&=
P_{A\mid U,s}\; P_{U\mid Z,s},
\label{eq:factor-A-new}
\\
P_{W\mid Z,s}
&=
P_{W\mid U}\; P_{U\mid Z,s},
\label{eq:factor-W-new}
\end{align}
where $P_{A\mid U,s} \in \mathbb R^{|\mathcal A|\times k}$,
$P_{W\mid U} \in \mathbb R^{|\mathcal W|\times k}$,
and $P_{U\mid Z,s} \in \mathbb R^{k\times |\mathcal Z|}$.

Next, by Assumption~\ref{assump:rank_discrete}, there exist coarsenings
$Z'_t$ of $Z_t$ and $W'_{t-1}$ of $W_{t-1}$, both taking values in $\{1,\dots,k\}$,
such that the matrix
\[
P_{W'\mid Z',s} \in \mathbb R^{k\times k}
\]
is square and invertible.
Restricting \eqref{eq:factor-W-new} to these coarsenings gives
\begin{align}
P_{W'\mid Z',s}
=
P_{W'\mid U}\; P_{U\mid Z',s}.
\label{eq:restricted-W}
\end{align}

Since $P_{W'\mid Z',s}$ is invertible, it follows that $P_{U\mid Z',s}$ is also invertible and
\begin{align}
P_{U\mid Z',s}^{-1}
=
P_{W'\mid Z',s}^{-1}\; P_{W'\mid U}.
\label{eq:U-inverse-new}
\end{align}

Now, restricting \eqref{eq:factor-A-new} to the same coarsening $Z'_t$ yields
\begin{align}
P_{A\mid Z',s}
=
P_{A\mid U,s}\; P_{U\mid Z',s}.
\label{eq:restricted-A}
\end{align}
Multiplying both sides of \eqref{eq:restricted-A} on the right by
$P_{U\mid Z',s}^{-1}$ and substituting \eqref{eq:U-inverse-new}, we obtain
\begin{align}
P_{A\mid U,s}
&=
P_{A\mid Z',s}\; P_{U\mid Z',s}^{-1} \notag \\
&=
P_{A\mid Z',s}\;
P_{W'\mid Z',s}^{-1}\;
P_{W'\mid U}.
\label{eq:A-given-U}
\end{align}

We now identify the interventional action distribution.
By consistency and the definition of the intervention $A_t^{(s)}$,
\begin{align*}
P\!\left(A_t^{(s)}\right)
&=
P_{A\mid U,s}\; P_U .
\end{align*}
Substituting \eqref{eq:A-given-U} into the above expression yields
\begin{align*}
P\!\left(A_t^{(s)}\right)
&=
P_{A\mid Z',s}\;
P_{W'\mid Z',s}^{-1}\;
P_{W'\mid U}\;
P_U .
\end{align*}
Since $P_{W'} = P_{W'\mid U}\, P_U$, we conclude that
\begin{align*}
P\!\left(A_t^{(s)}\right)
=
P_{A\mid Z',s}\;
\Big(P_{W'\mid Z',s}\Big)^{-1}\;
P_{W'} .
\end{align*}

Therefore, the interventional action distribution $P(A_t^{(s)})$ is identified.
Consequently,
\[
\pi_{\mathrm{opt}}(s)
\;\in\;
\arg\max_{a\in\mathcal A}
p\!\left(A_t^{(s)} = a\right)
\]
is identified for all $s$ (up to ties), completing the proof.

\subsection{Proof of Theorem \ref{thm:identification-continuous}}

Fix any $s$ in the support of $S_t$ and any action $a\in\mathcal A$.
Suppose that the function $h_a(w,s)$ satisfies the confounding bridge equation
\eqref{eq:bridge_eq}, that is,
\begin{align*}
p(A_t = a \mid S_{t-1}=z, S_t=s)
=
\int h_a(w,s)\, p(w \mid S_{t-1}=z, S_t=s)\, dw
\end{align*}
for all $z$ in the support of $S_{t-1}$.

Integrating both sides with respect to the latent variable $U_{t-1}$, we obtain
\begin{align*}
\int p(A_t = a, U_{t-1}=u \mid S_{t-1}, S_t=s)\, du
&=
\iint h_a(w,s)\,
p(w, U_{t-1}=u \mid S_{t-1}, S_t=s)\,
dw\, du .
\end{align*}

Rewriting both sides using the chain rule yields
\begin{align*}
\int p(A_t = a \mid U_{t-1}=u, S_{t-1}, S_t=s)\,
p(U_{t-1}=u \mid S_{t-1}, S_t=s)\, du \\
=
\iint h_a(w,s)\,
p(w \mid U_{t-1}=u, S_{t-1}, S_t=s)\,
p(U_{t-1}=u \mid S_{t-1}, S_t=s)\,
dw\, du .
\end{align*}

By Assumption~\ref{assump:proxy_ci}, conditional on $(U_{t-1}, S_t=s)$,
the action $A_t$ is independent of $S_{t-1}$ and the proxy $W_{t-1}$ is independent of $S_{t-1}$.
Therefore, the above equality simplifies to
\begin{align*}
\int p(A_t = a \mid U_{t-1}=u, S_t=s)\,
p(U_{t-1}=u \mid S_{t-1}, S_t=s)\, du \\
=
\iint h_a(w,s)\,
p(w \mid U_{t-1}=u)\,
p(U_{t-1}=u \mid S_{t-1}, S_t=s)\,
dw\, du .
\end{align*}
Then, Assumption~\ref{assump:completeness_cont} implies that
\begin{align*}
p(A_t = a \mid U_{t-1}=u, S_t=s)
=
\int h_a(w,s)\, p(w \mid U_{t-1}=u)\, dw
\end{align*}
By consistency and latent exchangeability,
\begin{align*}
p(A_t^{(s)} = a)
&=
\int p(A_t = a \mid U_{t-1}=u, S_t=s)\, p(U_{t-1}=u)\, du .
\end{align*}
Substituting the expression above yields
\begin{align*}
p(A_t^{(s)} = a)
&=
\int \left\{\int h_a(w,s)\, p(w \mid U_{t-1}=u)\, dw\right\}
p(U_{t-1}=u)\, du \\
&=
\int h_a(w,s)\,
\left\{\int p(w \mid U_{t-1}=u)\, p(U_{t-1}=u)\, du\right\}
dw \\
&=
\int h_a(w,s)\, p(w)\, dw .
\end{align*}
Therefore, the interventional action probability $p(A_t^{(s)}=a)$ is identified for all
$a\in\mathcal A$.
Consequently,
\[
\pi_{\mathrm{opt}}(s)
\;\in\;
\arg\max_{a\in\mathcal A} p(A_t^{(s)}=a)
\]
is identified for all $s$ (up to ties).
This concludes the proof.

\subsection{Proof of Proposition \ref{prop:rkhs_closed_form_causil}}
We first show that for any function $h$,
\begin{equation}
\label{eq:inner_sup_rewrite}
\sup_{q \in \mathcal Q}
\widehat{\mathbb E}\Big[
q(Z,S)\big\{ Y^{(a)} - h(W,S) \big\}
- q(Z,S)^2
\Big]
- \lambda_{\mathcal Q} \|q\|_{\mathcal Q}^2
=
\frac{1}{4}\,
\big\{\xi_N(h)\big\}^\top
K_{\mathcal Q}
\Big(
\frac{1}{N} K_{\mathcal Q} + \lambda_{\mathcal Q} I_N
\Big)^{-1}
\big\{\xi_N(h)\big\},
\end{equation}
where
\[
\xi_N(h)
=
\frac{1}{N}
\big(
Y^{(a)}_j - h(W_j,S_j)
\big)_{j=1}^N .
\]

To see this, we note that by the generalized representer theorem
\citep{scholkopf2001generalized},
the solution to the inner maximization problem over $q\in\mathcal Q$
admits the finite-dimensional representation
\[
q(z,s)
=
\sum_{j=1}^N \alpha_j\,
K_{\mathcal Q}\big((Z_j,S_j),(z,s)\big).
\]
Hence, the optimization over $q$ reduces to an optimization over the
coefficient vector
$\alpha = (\alpha_j)_{j=1}^N \in \mathbb R^N$.

Note that we can write
\[
q(Z_i,S_i)
=
\sum_{j=1}^N \alpha_j
K_{\mathcal Q}\big((Z_j,S_j),(Z_i,S_i)\big)
=
\big(K_{\mathcal Q}\alpha\big)_i .
\]
Hence,
\[
\sum_{i=1}^N q^2(Z_i,S_i)
=
\sum_{i=1}^N \big(K_{\mathcal Q}\alpha\big)_i^2
=
\|K_{\mathcal Q}\alpha\|_2^2
=
\alpha^\top K_{\mathcal Q}^2 \alpha .
\]

Therefore, we have
\begin{align*}
\widehat{\mathbb E}
\Big[
q(Z,S)\big\{ Y^{(a)} - h(W,S) \big\}
\Big]
&=
\frac{1}{N}
\sum_{i=1}^N
q(Z_i,S_i)\,
\big\{ Y_i^{(a)} - h(W_i,S_i) \big\} =
\alpha^\top
K_{\mathcal Q}\,
\{\xi_N(h)\}, \\[6pt]
\widehat{\mathbb E}\!\left[q^2(Z,S)\right]
&=
\frac{1}{N}
\sum_{i=1}^N q^2(Z_i,S_i)
=
\frac{1}{N}\,
\alpha^\top K_{\mathcal Q}^2 \alpha , \\[6pt]
\|q\|_{\mathcal Q}^2
&=
\alpha^\top K_{\mathcal Q}\alpha .
\end{align*}

Combining the above expressions, we obtain
\begin{align*}
\widehat{\mathbb E}
\Big[
q(Z,S)\big\{ Y^{(a)} - h(W,S) \big\}
- q^2(Z,S)
\Big]
- \lambda_{\mathcal Q} \|q\|_{\mathcal Q}^2
&=
\alpha^\top K_{\mathcal Q} \{\xi_N(h)\}
- \frac{1}{N}\alpha^\top K_{\mathcal Q}^2 \alpha
- \lambda_{\mathcal Q}\alpha^\top K_{\mathcal Q}\alpha \\
&=
\alpha^\top K_{\mathcal Q} \{\xi_N(h)\}
-
\alpha^\top
\Big(
\frac{1}{N} K_{\mathcal Q}^2
+ \lambda_{\mathcal Q} K_{\mathcal Q}
\Big)
\alpha .
\end{align*}

Let
\[
A
:=
\frac{1}{N} K_{\mathcal Q}^2
+ \lambda_{\mathcal Q} K_{\mathcal Q},
\qquad
b
:=
K_{\mathcal Q}\{\xi_N(h)\}.
\]
Then the objective function can be written as
\[
L(\alpha)
=
\alpha^\top b
-
\alpha^\top A \alpha .
\]
We therefore consider the maximization problem
\[
\max_{\alpha\in\mathbb R^N}
\;
\alpha^\top b
-
\alpha^\top A \alpha .
\]

Taking derivatives with respect to $\alpha$ and setting them to zero yields
\[
\nabla_\alpha L(\alpha)
=
b - 2A\alpha
= 0,
\]
which implies
\[
A\alpha
=
\frac{1}{2} b.
\]

Therefore, the optimal coefficients are given by
\begin{align*}
\alpha^*
&=
\frac{1}{2} A^{-1} b \\
&=
\frac{1}{2}
\Big(
\frac{1}{N} K_{\mathcal Q}^2
+ \lambda_{\mathcal Q} K_{\mathcal Q}
\Big)^{-1}
K_{\mathcal Q}
\{\xi_N(h)\} \\
&=
\frac{1}{2}
\Big(
\frac{1}{N} K_{\mathcal Q}
+ \lambda_{\mathcal Q} I_N
\Big)^{-1}
\{\xi_N(h)\},
\end{align*}
where $I_N$ denotes the $N\times N$ identity matrix.

Consequently, we have
\begin{equation}
\sup_{q \in \mathcal Q}
\widehat{\mathbb E}
\Big[
q(Z,S)\big\{ Y^{(a)} - h(W,S) \big\}
- q(Z,S)^2
\Big]
- \lambda_{\mathcal Q} \|q\|_{\mathcal Q}^2
=
\frac{1}{4}\,
\{\xi_N(h)\}^\top
K_{\mathcal Q}
\Big(
\frac{1}{N} K_{\mathcal Q}
+ \lambda_{\mathcal Q} I_N
\Big)^{-1}
\{\xi_N(h)\},
\end{equation}
where
\[
\xi_N(h)
=
\frac{1}{N}
\big(
Y^{(a)}_j - h(W_j,S_j)
\big)_{j=1}^N .
\]

Therefore, the outer minimization problem in~\eqref{eq:minimax_specific}
is reduced to
\begin{equation}
\widehat h
=
\arg\min_{h \in \mathcal H}
\;
\{\xi_N(h)\}^\top
\Gamma
\{\xi_N(h)\}
+
\lambda_{\mathcal H} \|h\|_{\mathcal H}^2,
\end{equation}
where
\[
\Gamma
=
\frac{1}{4}\,
K_{\mathcal Q}
\Big(
\frac{1}{N} K_{\mathcal Q}
+ \lambda_{\mathcal Q} I_N
\Big)^{-1}.
\]

We note that by the generalized representer theorem
\citep{scholkopf2001generalized},
the solution to this minimization problem admits the representation
\[
h(w,s)
=
\sum_{j=1}^N
\alpha_j\,
K_{\mathcal H}\big((W_j,S_j),(w,s)\big).
\]

Hence, only the coefficient vector
$\alpha = (\alpha_j)_{j=1}^N$
needs to be determined.

Recall that
\[
\xi_N(h)
=
\frac{1}{N}
\big(
Y^{(a)}_j - h(W_j,S_j)
\big)_{j=1}^N ,
\]
which can be written in vector form as
\[
\xi_N(h)
=
\frac{1}{N}
\Big(
y_a - K_{\mathcal H}\alpha
\Big),
\]
where $y_a := (Y^{(a)}_1,\dots,Y^{(a)}_N)^\top$.
Moreover,
\[
\|h\|_{\mathcal H}^2
=
\alpha^\top
K_{\mathcal H}
\alpha .
\]

Therefore, the outer optimization problem reduces to
\begin{align}
\min_{h \in \mathcal H}
\;
\{\xi_N(h)\}^\top
\Gamma
\{\xi_N(h)\}
+
\lambda_{\mathcal H} \|h\|_{\mathcal H}^2
&=
\min_{\alpha \in \mathbb R^N}
\;
\frac{1}{N^2}
\alpha^\top
K_{\mathcal H}
\Gamma
K_{\mathcal H}
\alpha
-
\frac{2}{N^2}
y_a^\top
\Gamma
K_{\mathcal H}
\alpha
+
\lambda_{\mathcal H}
\alpha^\top
K_{\mathcal H}
\alpha
+
c,
\end{align}
where
$c = \tfrac{1}{N^2} y_a^\top \Gamma y_a$
is a constant independent of $\alpha$.

This convex quadratic optimization problem admits the solution
\begin{equation}
\alpha^*
=
\Big(
K_{\mathcal H}
\Gamma
K_{\mathcal H}
+
N^2 \lambda_{\mathcal H}
K_{\mathcal H}
\Big)^{\dagger}
K_{\mathcal H}
\Gamma
y_a .
\end{equation}

\section{Numerical Specification of DGP Used in the Simulation Study}
\label{appendix:simu_setting}

In this subsection, we provide the full numerical specification of the data-generating process
used in the simulation studies.
The latent state, observed state, measurement variable, and action take values in
$\{0,\dots,K_U-1\}$, $\{0,\dots,K_S-1\}$, $\{0,\dots,K_W-1\}$, and $\{0,\dots,K_A-1\}$, respectively,
with $(K_U, K_S, K_W, K_A) = (4, 4, 4, 4)$.

The latent confounder $U_t \in \{0,\dots,K_U-1\}$ evolves according to
\[
\mathbb{P}(U_t = k \mid U_{t-1}, S_{t-1}, A_{t-1})
=
\operatorname{softmax}\!\Big(
\boldsymbol{\alpha}_0
+ \boldsymbol{A}_U U_{t-1}
+ \boldsymbol{A}_S S_{t-1}
+ \boldsymbol{A}_A A_{t-1}
\Big)_k,
\]
where $\boldsymbol{\alpha}_0 = \mathbf{0}$,
$\boldsymbol{A}_U = -0.5\,\boldsymbol{I}_{K_U}$,
$\boldsymbol{A}_S = 0.6\,\boldsymbol{I}_{K_U \times K_S}$,
and $\boldsymbol{A}_A = 0.15\,\boldsymbol{I}_{K_U \times K_A}$.

The observed state $S_t \in \{0,\dots,K_S-1\}$ is generated as
\[
\mathbb{P}(S_t = k \mid U_t, U_{t-1}, S_{t-1}, A_{t-1})
=
\operatorname{softmax}\!\Big(
\boldsymbol{\beta}_0
+ \boldsymbol{B}_U U_t
+ \boldsymbol{B}_{U^-} U_{t-1}
+ \boldsymbol{B}_S S_{t-1}
+ \boldsymbol{B}_A A_{t-1}
\Big)_k,
\]
with $\boldsymbol{\beta}_0 = \mathbf{0}$,
$\boldsymbol{B}_U = 0.15\,\boldsymbol{I}_{K_S \times K_U}$,
$\boldsymbol{B}_{U^-} = 0.1\,\boldsymbol{I}_{K_S \times K_U}$,
$\boldsymbol{B}_S = 0.35\,\boldsymbol{I}_{K_S}$,
and $\boldsymbol{B}_A = 0.15\,\boldsymbol{I}_{K_S \times K_A}$.

The measurement variable $W_t \in \{0,\dots,K_W-1\}$ is sampled according to
\[
\mathbb{P}(W_t = k \mid U_t)
=
\operatorname{softmax}\!\big(
\boldsymbol{\omega}_0 + \boldsymbol{\Omega}_U U_t
\big)_k,
\]
where $\boldsymbol{\omega}_0 = \mathbf{0}$ and
$\boldsymbol{\Omega}_U = 1.5\,\boldsymbol{I}_{K_W \times K_U}$.

The expert policy is deterministic and given by
\[
A_t
=
\pi_E(S_t, U_{t-1})
=
\arg\max_{k \in \{0,\dots,K_A-1\}}
\operatorname{softmax}\!\Big(
\boldsymbol{\gamma}_0
+ \boldsymbol{\Gamma}_S S_t
+ \boldsymbol{\Gamma}_U U_{t-1}
\Big)_k,
\]
with $\boldsymbol{\gamma}_0 = \mathbf{0}$,
$\boldsymbol{\Gamma}_S = 0.15\,\boldsymbol{I}_{K_A \times K_S}$,
and $\boldsymbol{\Gamma}_U = 1.5\,\boldsymbol{I}_{K_A \times K_U}$.

\section{Data Preprocessing Details \label{appendix:data_preprocessing}}
\medskip
\noindent\textbf{Variables Description.}
We focus on a Vital signs routinely collected ICU measurements that capture
cardiovascular, respiratory, and hemodynamic status.
The observed state at time $t$ is represented by a vector
$S_t = (\text{MAP}_t, \text{HR}_t, \text{DBP}_t, \text{SBP}_t, \text{O2Sat}_t, \text{Resp}_t)$,
while lactate, $U_t = (\text{Lactate}_t)$, is treated as a latent state variable.
Table~\ref{tab:variables} summarizes the clinical meaning of the variables used in our analysis.

\begin{table}[H]
\centering
\caption{Clinical variables used in the analysis.}
\label{tab:variables}
\begin{tabular}{ll}
\hline
\textbf{Variable} & \textbf{Description} \\
\hline
MAP   & Mean arterial pressure (mm Hg)\\
HR    & Heart rate (beats per minute)\\
DBP   & Diastolic blood pressure (mm Hg)\\
SBP   & Systolic blood pressure (mm Hg)\\
O2Sat & Pulse oximetry (\%)\\
Resp  & Respiration rate (breaths per minute)\\
Lactate & Blood lactate concentration (mg/dL)\\
\hline
\end{tabular}
\end{table}

\textbf{Missing data imputation.} We perform imputation at the patient level using ICU time (ICULOS) as the temporal index.
Patients for whom respiration rate (Resp) is entirely unobserved throughout the ICU stay are excluded, as no data-driven imputation is possible in such cases.
For the remaining patients, missing values in Lactate and vital signs are imputed using B-spline interpolation. Outside the observed time range, forward and backward filling is applied to obtain complete trajectories.

\textbf{Discretization of state and latent variables.}
After imputing missing values, all variables are discretized via quantile-based method based on their clinical interpretation \citep{kusumoto20192018, yu2023expert, lan2025admission}. For variables with well-established clinical interpretation and risk stratification,
namely mean arterial pressure (MAP), heart rate (HR), and systolic blood pressure (SBP),
we adopt a clinically motivated three-level discretization.
Specifically, MAP is discretized into low perfusion $(=0)$, normal $(=1)$, and elevated $(=2)$ regimes;
HR into bradycardia $(=0)$, normal $(=1)$, and tachycardia $(=2)$;
and SBP into hypotension $(=0)$, normotension $(=1)$, and hypertension $(=2)$.
These categories correspond to standard clinical thresholds and represent qualitatively distinct physiological states that are known to influence treatment decisions.

For the remaining vital signs—diastolic blood pressure (DBP), oxygen saturation (O2Sat), and respiration rate (Resp)—whose empirical distributions exhibit limited variability and strong concentration around the normal range,
we apply a binary discretization based on empirical quantiles.
Concretely, each of these variables is discretized into a lower $(=0)$ and higher $(=1)$ category. Finally, Lactate is discretized into two categories representing lower $(=0)$ and higher $(=1)$ severity levels.

\textbf{Generation of expert actions and proxy measurements.}
Let $S_t = (S_{t,1},\ldots,S_{t,d})$ denote the discretized observed state at time $t$,
where $d=6$ corresponds to the selected vital signs.
The components of $S_t$ take discrete values, with
$S_{t,j} \in \{0,1,2\}$ for MAP, HR, and SBP, and $S_{t,j} \in \{0,1\}$ for DBP, O2Sat, and Resp.
Let $U_t \in \{0,1\}$ denote the discretized latent severity state.

The expert policy is deterministic and depends on the current observed state
and the previous latent severity.
Specifically, the expert action is given by
\[
A_t
=
\pi_E(S_t, U_{t-1})
=
\arg\max_{k \in \{0,1,2\}}
\operatorname{softmax}\!\Big(
\boldsymbol{\gamma}_0
+
\boldsymbol{\Gamma}_S S_t
+
\boldsymbol{\Gamma}_U U_{t-1}
\Big)_k,
\]
where $\boldsymbol{\gamma}_0 = \mathbf{0} \in \mathbb{R}^3$,
$\boldsymbol{\Gamma}_S \in \mathbb{R}^{3 \times d}$,
and $\boldsymbol{\Gamma}_U \in \mathbb{R}^{3 \times 1}$.

We further consider two proxy measurements of the latent state.
For each proxy $W^{(j)}_t$, $j=1,2$, observations are generated as
\[
W^{(j)}_t
\sim
\text{Categorical}\!\left(
\operatorname{softmax}\!\big(
\boldsymbol{\omega}^{(j)}_0
+
\boldsymbol{\Omega}^{(j)}_U U_t
\big)
\right),
\qquad
W^{(j)}_t \in \{0,1,2\},
\]
where $\boldsymbol{\omega}^{(j)}_0 \in \mathbb{R}^3$ and
$\boldsymbol{\Omega}^{(j)}_U \in \mathbb{R}^{3 \times 1}$.

Next, to evaluate performance of different polices under different types of distributional shift,
similar to simulation studies in Section \ref{sec:experiments}, we construct two controlled shift scenarios.

\medskip
\noindent\textbf{Shift on noisy state observations.}
We consider a measurement mechanism shift that alters the conditional distribution
of the proxy variable $W_t$ given the latent state $U_t$,
while leaving the latent process, observed states, and expert actions unchanged.
In the baseline setting, the proxy is generated according to
\[
\mathbb{P}(W_t = k \mid U_t)
=
\operatorname{softmax}\!\left(
\boldsymbol{\omega}_0 + \boldsymbol{\Omega}_U U_t
\right)_k.
\]
To induce a measurement shift, we modify the proxy mechanism for observations
appearing after a fixed cutoff index $t_0 = 4000$,
which corresponds to the beginning of the test set.
Specifically, for all $t \ge t_0$, the proxy-generating parameters are sign-flipped:
\[
\mathbb{P}_{\text{shift}}(W_t = k \mid U_t)
=
\operatorname{softmax}\!\left(
-\boldsymbol{\omega}_0 - \boldsymbol{\Omega}_U U_t
\right)_k.
\]
This intervention changes the relationship between $W_t$ and $U_t$
without affecting the expert policy, latent state dynamics,
or the observed state variables.

\medskip
\noindent\textbf{Population shift.}
We additionally consider a population shift implemented through selective sampling.
Specifically, we operate on the test set based on temporal ordering,
consisting of observations with index $t \ge t_0 = 4000$.
From this pool, we construct the final test set by jointly imposing two
clinically motivated criteria:
(i) the lactate level lies in the top $10\%$ of its empirical distribution
(corresponding to $\text{Lactate} \ge 10.442$ which is the $90\%$ quantile),
and (ii) the patient has remained in the ICU for at least $12$ hours.
The resulting test set therefore concentrates on a subpopulation of more severely ill,
late-stage ICU patients.

\end{document}